\pdfoutput=1

\documentclass[journal]{IEEEtran}
\usepackage{cite}
\usepackage{bm}
\usepackage{algorithm}
\usepackage{algorithmic}
\usepackage{graphicx}
\usepackage{lineno,hyperref}
\usepackage{subfigure}
\usepackage{amsmath}
\usepackage{pifont}
\usepackage{epstopdf}
\usepackage{color}
\usepackage{multirow}
\usepackage{amsthm,amsmath,amsfonts,amssymb,bm}
\usepackage{array}
\usepackage{booktabs}
\usepackage{makecell}
\usepackage[all]{xy}

\modulolinenumbers[5]

\ifCLASSINFOpdf
\else
\fi
\begin{document}
%
\title{Two-level Graph Neural Network}
%
%
%

\author{Xing~Ai, Chengyu~Sun, Zhihong~Zhang$^{\ast}$, and~Edwin~R~Hancock,~\IEEEmembership{Fellow,~IEEE}
    \IEEEcompsocitemizethanks{
        \IEEEcompsocthanksitem Xing Ai and Chengyu Sun are with School of Informatics, Xiamen University, Xiamen, Fujian, China.
        \protect\\ E-mail: 24320191152507@stu.xmu.edu.cn, 30920201153942@stu.xmu.edu.cn
        \IEEEcompsocthanksitem Corresponding author: Zhihong Zhang is with School of Informatics, Xiamen University, Xiamen, Fujian, China.
        \protect\\ E-mail: zhihong@xmu.edu.cn
        \IEEEcompsocthanksitem Edwin R. Hancock is with University of York, York, UK.
        \protect\\ E-mail: edwin.hancock@york.ac.uk}}

%
%

\markboth{Journal of \LaTeX\ Class Files,~Vol.~14, No.~8, August~2015}%
{Shell \MakeLowercase{\textit{et al.}}: Bare Demo of IEEEtran.cls for IEEE Journals}
%



\maketitle

\begin{abstract}
    Graph Neural Networks (GNNs) are recently proposed neural network structures for the processing of graph-structured data. Due to their employed neighbor aggregation strategy, existing GNNs focus on capturing node-level information and neglect high-level information. Existing GNNs therefore suffer from representational limitations caused by the Local Permutation Invariance (LPI) problem. To overcome these limitations and enrich the features captured by GNNs, we propose a novel GNN framework, referred to as the Two-level GNN (TL-GNN). This merges subgraph-level information with node-level information. Moreover, we provide a mathematical analysis of the LPI problem which demonstrates that subgraph-level information is beneficial to overcoming the problems associated with LPI. A subgraph counting method based on the dynamic programming algorithm is also proposed, and this has time complexity is $O(n^3)$, $n$ is the number of nodes of a graph. Experiments show that TL-GNN outperforms existing GNNs and achieves state-of-the-art performance. 
\end{abstract}

\begin{IEEEkeywords}
Graph representation, Graph neural networks, Local permutation invariance, Attention mechanism. 
\end{IEEEkeywords}

%
\IEEEpeerreviewmaketitle

\section{Introduction}
%
%
%
%
\IEEEPARstart{G}{raph} Neural Networks (GNNs) have attracted increasing interest in graph-structured data such as social networks, recommender systems, bioinformatics and combinatorial optimization. Scarselli et al.\cite{scarselli2008graph} first introduced the concept of the GNN by extending recursive neural networks. Veličković et al.\cite{velickovic2018graph} proposed the Graph Attention Network (GAT), which leverages masked self-attentional layers to address the shortcomings of prior methods based on graph convolutions. Xu et al. \cite{xu2018powerful} present a theoretical framework 
to analyze the representational capability of GNNs, and develop a simple neural architecture referred to as the Graph Isomorphism Network (GIN). 

Although the existing neighborhood aggregation strategy used in GNNs is relatively efficient from the viewpoint of graph isomorphism classification, recent studies\cite{garg2020generalization}\cite{sato2019approximation}\cite{klicpera2020directional} show that such a procedure brings some inherent problems. Namely, most existing GNNs suffer from local permutation invariance (LPI), which leads them to confuse specific structures. In fact, invariance is very common in many learning tasks. Data can produce identical embeddings in a reduced low-dimensional space after symmetric transformations or rotations are applied\cite{krizhevsky2012imagenet,weiler20183d}. As for graph-structured data, Garg et al.\cite{garg2020generalization} have found that existing GNNs have representational limitations caused by the translation of graph-structured data. To eliminate such an effect, Sato et al.\cite{sato2019approximation} have exploited a local port ordering of nodes referred to as the Consistent Port Numbering GNN (CPNGNN). Moreover, Klicpera et al.\cite{klicpera2020directional} have proposed DimeNet, which is a directional message passing algorithm introduced in the context of molecular graphs. However, Garg V et al.\cite{garg2020generalization} prove that all existing GNN variants have representational limits caused by LPI, and propose a novel graph-theoretic formalism.

In the meantime, studies show that complex networks can be succinctly described using graph substructures (also referred to as subgraphs, graphlets, or motifs). Subgraph methods have been well-studied and widely used in chemistry\cite{duvenaud2015convolutional}, biology\cite{koyuturk2004efficient}, and social network graph tasks\cite{jiang2010finding}. For example, specific patterns of atoms or modes of interaction can be discovered by identifying specific subgraph topologies. Bai et al.\cite{bai2020ripple} propose a general subgraph-based training framework referred to as Ripple Walk Training (RWT). This can not only accelerate the training speed on large graphs but also solve problems associated with the memory bottleneck. Emily et al.\cite{alsentzer2020subgraph} propose SUBGNN to propagate neural messages between the subgraph components and randomly sampled anchor patches. These methods extract node features and subgraph features separately. Moreover, they characterize only the number of different subgraphs, which ignore the learning of their representation. 

For the sake of the above mentioned problems, we propose a novel model which merges subgraph-level information into the node-level representation. First, we merge subgraph-level information at the node-level to enrich the features. And secondly, we theoretically verify the model to demonstrate its performance with real-world datasets. The results show that our approach is significantly more effective than state-of-the-art baselines. Our main contributions are summarized as follows:

\begin{itemize}
  \item [1.] We propose a novel GNN approach, the Two-level GNN (TL-GNN), which captures both microscopic (small scale) and macroscopic (large scale) structural information simultaneously and thus enriches the representation of a graph.
  \item [2.] We provide a mathematical definition and analysis of the effects of LPI on GNNs. Furthermore, we prove that subgraph-level information offers benefits in overcoming these limitations. 
  \item [3.] We verify our method on seven different benchmarks and a synthetic dataset. The results show that TL-GNN is more powerful than existing GNN's. 
  \item [4.] A subgraph counting method based on dynamic programming is also proposed. The time complexity and space complexity of this algorithm are $O(n^3)$ and $O(n^3)$ respectively, where $n$ is the number of nodes in the graph.
\end{itemize}

The remainder of this paper is organized as follows. Section.~\ref{Related Work} provides an overview of the related work. Section.~\ref{Proposed Method} introduces the proposed method, including theoretically proving the capacity of our model to solve the LPI problem. Section.~\ref{Experiments} describes our experimental setting and demonstrate empirically the performance of TL-GNN. Finally, Section.~\ref{Conclusion} concludes the paper and offers directions for future work.

\section{Related Work}
\label{Related Work}

GNNs have achieved state-of-the-art results on graph classification, link prediction and semi-supervised node classification.However, recent studies\cite{garg2020generalization}\cite{sato2019approximation}\cite{klicpera2020directional} have demonstrated one of the severe representational limitations of GNNs, namely Local Permutation Invariance (LPI). In this section, we will briefly review these interrelated topics.

\subsection{Graph Neural Networks.} Graph Neural Networks (GNNs) have proved to be an effective machine learning tool for non-Euclidean structure data for several years. Since the GNN was first presented in \cite{scarselli2008graph}, a set of more advanced approaches have been proposed, including but not limited to GraphSAGE \cite{NIPS2017_6703}, Graph Attention Networks (GAT) \cite{velickovic2018graph}, Graph Isomorphism Network (GIN) \cite{xu2018powerful}, edGNN \cite{jaume2019edgnn}. These methods learn local structural information by recursively aggregating neighbor representations.


In a macroscopic view, the above models follow the same pattern. For each node $v\in V$ within a graph $G=(V,E)$, GNNS capture $k$-hop neighbor information $h_{\mathcal{N}(v)}^{k}$, and then learn a representational vector $h_{v}^{k}$ after $k$ layers of processing. On this basis, tasks such as graph classification can be accomplished. In fact, the critical difference between GNN variants is how they design and implement the neighbor aggregation function. Xu et al.\cite{xu2018powerful} summarized some of the most common GNN approaches and proposed a general framework referred to as the Graph Isomorphism Network (GIN). The GIN approach defines the above steps as three related functions, namely AGGREGATE (AGG), COMBINE (COM), and READOUT (READ), 

\begin{equation}
    \begin{cases}
    h_{\mathcal{N}(v)}^{k}=AGG(\{h_\mu^{(k-1)},\forall\mu\in\mathcal{N}(v)\}),\\
    \\
    h_{v}^{k}=COM(h_v^{k-1},h_{\mathcal{N}(v)}^{k}),\\
    \\
    h_G=READ(\{ h^k_v|v\in G \} ),
    \end{cases}\label{GNN_form}
\end{equation}

The initialization is $h_{v}^0=X_v$, and $X_v$ represents the initial features of the nodes. The quantity $\mathcal{N}(v)$ represents the set of nodes adjacent to $v$ and $h_G$ is the graph representation vector. 

Xu et al.\cite{xu2018powerful} indicate that what makes GNN so powerful is the injective aggregation strategy, which maps different nodes to different representational units. They also demonstrate that when the above three functions are all injective functions, for example a sum, then the GNN can be as powerful as the WL test\cite{2010Weisfeiler} on the graph isomorphism problem.



\begin{figure*}
    \centering
    \includegraphics[scale=0.5]{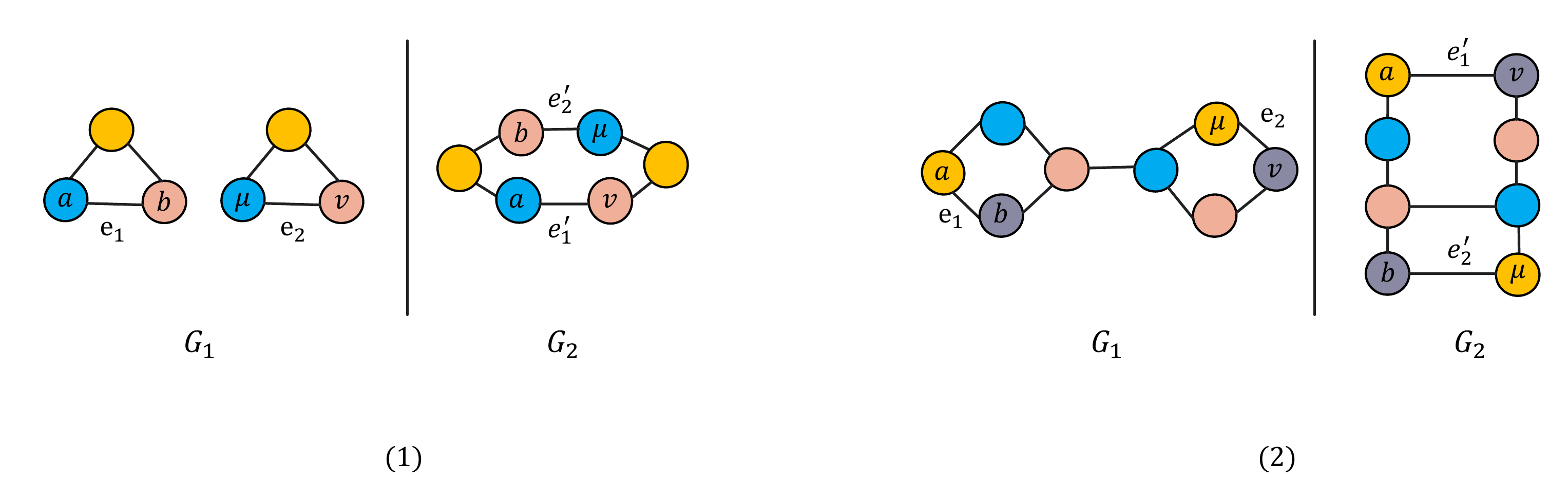}
    \caption{Two LPI examples, $G_1$ and $G_2$ have identical node-level information but different structures. In other words, they are non-isomorphic graphs. The nodes of the same colour have the same features. The numbers of nodes and edges number together with the node features are identical. The only difference between them is that a pair of edges of $G_1$: $e_1=(a,b)$, $e_2=(\mu,v)$, change their nodes and transform the edges of $G_2$: so that $e_{1}^{'}=(a,v)$, $e_{2}^{'}\prime=(b,\mu)$. }
    \label{LPI example}
\end{figure*}

\subsection{Local Permutation Invariance}
Even though the precise form of the aggregation or combination strategy varies across different GNN architectures, most share the same neighborhood aggregation concept. This characteristic leads to an underlying graph isomorphism problem, the so-called local permutation invariance (LPI). {\color{red}{A more common LPI example in the real world is edge rewiring\cite{zhou2014memetic}\cite{rong2018heuristic}. For a graph $G=(V,E)$, edge rewiring operation alters the graph structure and leads to a new graph $G^{'}=(V^{'},E^{'})$ by exchanging a pair of edges, for instance removing edges $e_{AB}, e_{CD}$ between nodes $A,B,C,D$ and adding edges $e_{AC}, e_{BD}$. After changing, $G$ and $G^{'}$ are non-isomorphic graphs but have identical representation in GNNs: $h_{G}=h_{G^{'}}$.}} 

Garg et al.\cite{garg2020generalization} analyze specific cases of LPI for the GNNs aggregation function and provide generalization bounds for message passing in GNNs. As shown in Fig.~\ref{LPI example}, the graphs are obviously non-isomorphic, but their node-level characteristics are identical. Unfortunately, existing GNNs focus on extracting node-level information and neglect high-level information, so they suffer representational limitations. Recent studies have attempted to overcome these limitations by providing additional information to the nodes. Consistent Port Numbering GNN (CPNGNN)\cite{sato2019approximation} assigns port numbers to nodes and treats their neighbors differently. Klicpera et al.\cite{klicpera2020directional} propose DimNet for molecular graphs. This embeds whole atoms using a collection of edge embeddings and takes advantage of directional information by modifying messages based on their angle. However, Garg et al.\cite{garg2020generalization} demonstrate that there are some graphs that both CPNGNN and DimeNet can not distinguish. This means that these approaches fail to overcome the representational limitations caused by LPI.

\begin{figure}[ht]
    \centering
    \includegraphics[scale=0.45]{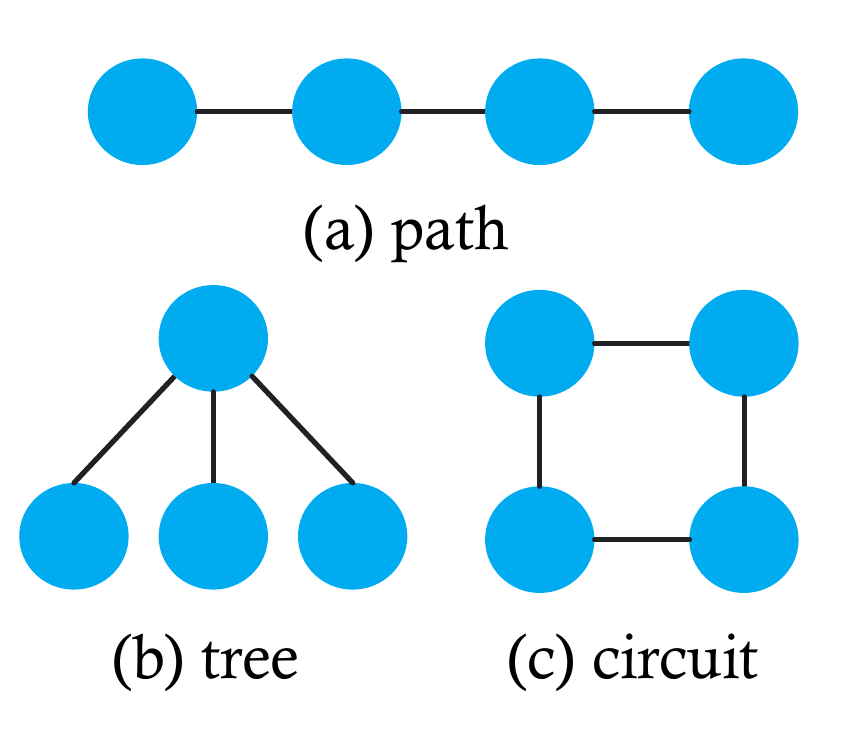}
    \caption{Subgraphs example}
    \label{subgraphs}
\end{figure}

\subsection{Subgraph Methods.} 
Subgraphs can be regarded as the basic structural elements or building blocks of the larger graph, including paths\cite{borgwardt2005shortest} and subtrees\cite{shervashidze2009efficient}. Subgraph frequency was studied as a local feature in social networks by Ugander et al. \cite{ugander2013subgraph}, who discovered that it can provide unique insights for recognizing both social organization and graph structure in very large networks. Ugender et al.\cite{ugander2013subgraph} propose a novel graph homomorphism algorithm based on subgraph frequencies. They define a coordinate system that is beneficial both to the representation and understanding of large sets of dense graphs. Grochow et al.\cite{grochow2007network} introduce a symmetry-breaking technique into the motif (subgraph) discovery process and develop a novel motif discovery algorithm that can achieve an exponential speed-up. More recently, significant effort has been expended in the design of subgraph-based GNNS for graph classification. The role of substructures has been explored empirically by Ying et al.\cite{ying2019gnnexplainer} and used to interpret the predictions made by GNNs.

In this paper, we propose an architecture which we refer to as the Two-level Graph Neural Network (TL-GNN), which is designed to make the association between the neighbor node and subgraph structural information. Our model mitigates the negative impacts of the LPI problem and is verified to be efficient on real-world datasets.

\section{Proposed Method}
\label{Proposed Method}
In order to extract subgraph-level information from a graph, we first count all subgraphs within a certain size range of a graph and then generate a new graph, namely generated graph, whose nodes (or supernodes) represent subgraphs in the original graph. 
We develop a novel subgraph counting algorithm for this purpose. After the generated graph is to hand, we develop a novel GNN framework, that significantly extends the existing GNN framework. We also develop two key operators for the novel framework that facilitate the effective merging of subgraph and node information. This Section will introduce the proposed subgraph counting algorithm and the new GNN framework.

\subsection{Constructing the Generated Graph}
\label{Constructing Generated Graph}
Consider an undirected graph $G=(V,E)$, where $V$ and $E\subseteq (V\times V)$ respectively denote the set of nodes and the set of edges. The element $(v_i, v_j)$ in $E$ is an unordered pair of nodes $v_i$ and $v_j$, $i,j=1,2,3...N$, where N is the number of nodes in the graph, i.e. the size of the network. 

\begin{algorithm}[tb]
\caption{Tree counting}
\begin{algorithmic}[1]
    \STATE{\bfseries Input}: input original graph $G(V,E)$ and its adjacency matrix $A$, hyper parameter $tree\_threshold$.
    \STATE{\bfseries Output}: output subgraph set $S$
    
     \STATE Initialize subgraphset $S$, tree set $Tree$ , neighbor set $\{N(v)|t=0,1,2...2^T,\forall v\in V \} $, and path from $v$ to $\mu : \{P_{v}(\mu)\vert\forall v,\mu\in V\} $
    
    \FOR{$v \in V$}
        \FOR{$\mu \in V$}
            \IF {$A_{v\mu}=1$}
                \STATE add $\mu$ into $N(v)$
        
                    add $v$ into $N(\mu)$
        
                    $P_v(\mu)\gets {v}$
        
                    $P_\mu(v)\gets {\mu}$
            \ENDIF
        \ENDFOR
        \IF{$|N(v)|>tree\_threshold$}
            \STATE add $(N(v)+[v])$ into $Tree$
        \ENDIF
    \ENDFOR
    
    \STATE add $Tree$ into $S$
    \STATE return $S$
\end{algorithmic}
\label{Tree counting}
\end{algorithm}

\subsubsection{Subgraph Counting Algorithm }
We aim to identify three different types of subgraph structures, namely trees, paths, and circuits (cycles). These represent the different basic classes of subgraph structure, and structures such as triangles or quadrilaterals can be regarded as different specific cases as shown in Fig.\ref{subgraphs}. Finding all subgraphs within a graph is an NP-hard problem. To implement our method, it is unnecessary to find all subgraphs. For each node in turn, we identify all subgraphs that are contained within its $D$-hop neighbors. We design a dynamic programming algorithm to achieve this goal. Our algorithm consists of three steps for  tree, path, and circuit location. Firstly, we store the neighbor set of each node $v\in V$ and select those tree-shaped subgraphs which have more than three nodes ($tree\_threshold=3$). A 3-node tree or a 2-node tree is essentially a 3-node path or a 2-node path. This step is shown in Algorithm.~\ref{Tree counting}. Secondly, we find all path-shaped and circuit-shaped subgraphs based on the dynamic programming algorithm. This step is realized using Algorithm.~\ref{Path and circuit counting} and Algorithm.~\ref{Path and circuit sifting}. If the node $\mu$ is one of the $2^{d}$-hop neighbors of the node $v$ and the node $a$ is one of the $i$-hop neighbors of the node $\mu$ ($i\leq2^{d}$), then the node $a$ is one of the $(2^{d}+i)$-hop neighbors of the node $v$. After Algorithm.~\ref{Tree counting} locates the 1-hop neighbors of each node, Algorithm.~\ref{Path and circuit counting} and Algorithm.~\ref{Path and circuit sifting} can locate the 2-hop neighbors of each node by two sequential 1-hop searches. After locating the 2-hop neighbors of each node, the 3-hop neighbors and 4-hop neighbors of each node can be found. Besides, we store all nodes on the path from the node $v$ to the node $\mu$. If the node $\mu$'s $i$-hop neighbor $a$ is not on the path from the node $v$ to the node $\mu$ and the node $v$ has a path to the node $a$, a circuit can be found. Otherwise, a path can be obtained. 

\begin{algorithm}[tb]
\caption{Path and circuit counting}
\begin{algorithmic}[1]
    \STATE{\bfseries Input}: input original graph $G(V,E)$, depth $D$, subgraph set $S$ and $d$-hop adjacency matrix set $\{A^d\in \mathbb{R}^{\vert V\vert\times \vert V\vert} \vert d=2...2^D\} $ and path from $v$ to $\mu : \{P_{v}(\mu)\vert\forall v,\mu\in V\} $
    \STATE{\bfseries Output}: output subgraph set $S$
    
    \FOR{$d=0,1,2...D$}
        \FOR{$v\in V$}
            \FOR{$\mu\in V$}
                \FOR{$a\in V$}
                    \STATE $Circuit, Path$=Algorithm3
                \ENDFOR
            \ENDFOR
        \ENDFOR
    \ENDFOR
\STATE add $Circuit, Path$ to $S$
\STATE return $S$
\end{algorithmic}
\label{Path and circuit counting}
\end{algorithm}

{\color{red}{The main advantage of the proposed algorithm is low complexity. Early subgraph counting algorithms\cite{kloks2000finding}\cite{wernicke2006efficient}\cite{shervashidze2009efficient} required exponential time complexity. The current best-known algorithm \cite{eisenbrand2004complexity} for exact subgraph counting, which requires $O(n^{\frac{\omega D}{3}})$, where $\omega$ and $D$ are exponent of fast matrix multiplication and nodes number of subgraphs. Due to $\omega$ can be neglected, the time complexity of this algorithm is $O(n^D)$. However, the proposed method only requires $O(n^{3})$. As for the proposed methods, the time complexity of Algorithm.~\ref{Tree counting} is $O(n^2)$, $n=\vert V\vert$. The time complexity of Algorithm.~\ref{Path and circuit counting} and Algorithm.~\ref{Path and circuit sifting} is $O(2^{D}n^{3})$. Due to the fact that $D$ is a scalar no larger than 10, the  factor of $2^D$ can be neglected. The time complexity of the proposed method is therefore $O(n^{3})$. }}

As for space complexity, space complexities of $P_u(v)$ and $A_d$ are $O(n^3)$ and $O(2^{D}n^{2})$ respectively. As a result the space complexity of the proposed subgraph counting method is $O(n^3)$.

\subsubsection{Generating Graphs}
\label{Generating Graphs}
\begin{figure*}[ht]
    \centering
    \includegraphics[scale=0.35]{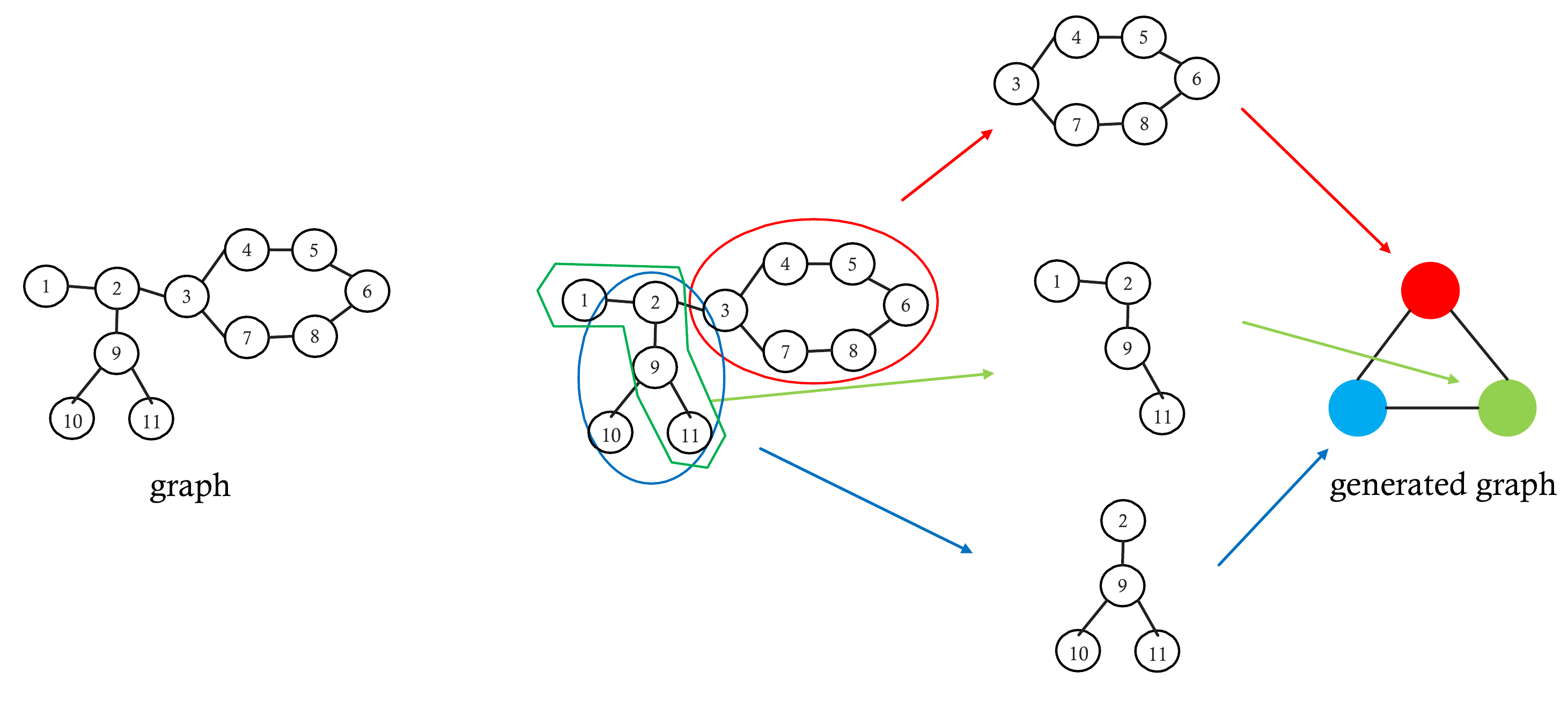}
    \caption{Example of the graph generation process. The red supernode corresponds to a 6-node circuit in the graph. The green and blue supernodes correspond to a 4-node path and a 4-node tree in the graph respectively.}
    \label{example generated graph}
\end{figure*}

After the subgraph counting is complete, we generate a new graph to represent the subgraph relationships. Given a network $G(V,E)$, the generated graph $G^*(V^*,E^*)$ consists of the set of supernodes representing the detected subgraphs $V^*$ and the set of edges $E^*\subseteq (V^*\times V^*)$ representing the relationships between them. Two subgraphs are connected if they share common nodes or links in the original network. The features associated with the supernodes are represented by a two-dimensional vector, whose components are the node counts
and the subgraph type respectively. Fig. \ref{example generated graph} provides an example.

\begin{algorithm}[H]
\caption{Path and subgraph sifting}
\begin{algorithmic}[1]
    \STATE{\bfseries Input}: input original graph $G(V,E)$, adjacency matrix $A$, subgraph set $S$, path from $v$ to $\mu : \{P_{v}(\mu)\vert\forall v,\mu\in V\} $ and $\{A^d|d=2...2^D\} $
    \STATE{\bfseries Output}: output circuit set $Circuit$ and path set $Path$
    
    \FOR{$i=1,2,...,2^{d}$}
        \IF{$A_{v \mu}^{(2^d)}=1$ and $A_{\mu a}^{(i)}=1$} 
            \IF {$a \notin P_{\mu}(v)$ and $P_{v}(a)\neq\varnothing$}
                \STATE add $P_{v}(\mu)+P_{\mu}(a)+P_{a}(v)$ to $Circuit$
            \ENDIF
            \IF {$a \notin P_{\mu v}$ and $P_{va}=\varnothing$}
                \STATE $A_{av}^{(2^d+i)}=1$
                \STATE $A_{va}^{(2^d+i)}=1$
                \STATE $P_{a}(v)=P_{a}(\mu)+P_{\mu}(v)$
                \STATE $P_{v}(a)=P_{v}(\mu)+P_{\mu}(a)$
                \STATE add $P_{v}(a)$ to $Path$
            \ENDIF
        \ENDIF
    \ENDFOR
\STATE return $Circuit$, $Path$
\end{algorithmic}
\label{Path and circuit sifting}
\end{algorithm}

We also record the appearances of each node in $V^*$ by constructing a transformation matrix $T\in\mathbb{R}^{N\times M}$, $\vert V\vert=N, \vert V^*\vert=M$. The transformation matrix $T$ indicates the correspondences between nodes and subgraphs (i.e. supernodes), i.e. which supernodes subsume each node. The elements of the transformation matrix are defined as follows:

\begin{equation}
    T_{ij}=
    \begin{cases}
    1, \quad if\ node\ i \ in \ subgraph \ j\\
    0, \quad else
    \end{cases}
\end{equation}

\subsection{The LPI problem and subgraphs}

{\color{red}{Garg et al.\cite{garg2020generalization} indicate the limitations of GNNs caused by Local Permutation Invariance (LPI). Specifically, changing a pair of edges in a graph leads to structure changing, while node-level information of the graph is maintained. The existing aggregation strategy, which extracts node-level information only is unable to distinguish structure change. Garg et al.\cite{garg2020generalization} provide several example structures which can not be distinguished by existing methods, as illustrated in Fig.~\ref{LPI example}. Specifically, two non-isomorphic graphs, $G$ and $G^{'}$ have identical node-level information which confuses GNNs to distinguish them. $G$ and $G^{'}$ can be transformed into each other by exchanging a pair of edges. It means their only difference is endpoints of the pair of edges.}}

However, Garg et al.\cite{garg2020generalization} provide examples only, without giving mathematical definitions for the ambiguities encountered.

Without loss of generality, we firstly indicate the general characteristics of graphs that existing GNNs can not distinguish due to the LPI ambiguity. Secondly, we demonstrate that for those graphs that GNNs can not distinguish due to ambiguities, our generated graphs behave in a different and useful manner. Finally, we demonstrate mathematically the effectiveness of our method in Section~\ref{Framework of TL-GNN}.

These results concerning LPI lead us to propose a new neighborhood aggregation strategy for GNNs. While the existing neighborhood aggregation strategy is effective, it sometimes fails to distinguish non-isomorphic graphs. To demonstrate this we give a definition of Permutation Non-isomorphic Graphs(PNG):

\textbf{Definition 1.} Permutation Non-isomorphic Graphs (PNG) are non-isomorphic graphs that have the same node set and the same node features but swap the nodes of two edges.

Assume $G=\{V,E\}$ and $G^{'}=\{V^{'},E^{'}\}$ are PNG. $v\in V, v^{'}\in V^{'}, e_1,e_2 \in E, e^{'}_1,e^{'}_2 \in E^{'}$, then they must satisfy the following conditions:

\begin{equation}
    \left\{
        \begin{array}{cc}
             \vert V\vert= \vert V^{'}\vert,& (a)\\
             
             \vert E\vert= \vert E^{'}\vert,& (b)\\
             
             h_v^{0}=h_{v^{'}}^{0} \quad \forall v\in V, & (c)\\
             
             N(v)=N(v^{'}), \quad \forall v \in V, & (d)\\
             
             e_1=\{a,b\}, e_2=\{i,j\},e^{'}_1=\{a^{'},j^{'}\},e^{'}_2=\{i^{'},b^{'}\}, \quad & (e)\\
             
        \end{array}
    \right. \label{coditions}
\end{equation}

The above conditions indicate that the only difference between $G=\{V,E\}$ and $G^{'}=\{V^{'},E^{'}\}$ is a single pair of edges. The remaining characteristics of $G=\{V,E\}$ and $G^{'}=\{V^{'},E^{'}\}$ are identical.

Due to LPI, the GNNs based neighbor aggregation strategy can not distinguish permutation non-isomorphic graphs, which we state in Theorem 2:

\textbf{Theorem 2.} GNNs based on Eq\eqref{GNN_form} can not distinguish permutation non-isomorphic graphs.

We give a proof of this theorem as follows. For two permutation non-isomorphic graphs $G=(V,E)$ and $G^{'}=(V^{'},E^{'})$, their neighbourhood aggregation from the 0-th GNN layer are:

\begin{equation}
    h_{\mathcal{N}(v)}^{0}=AGG(\{h_\mu^{(0)}\vert\mu\in\mathcal{N}(v)\}), \forall v\in V,
\end{equation}

\begin{equation}
    h_{\mathcal{N}(v^{'})}^{0}=AGG(\{h_\mu^{(0)}\vert\mu\in\mathcal{N}(v^{'})\}), \forall v^{'}\in V^{'}.
\end{equation}

Due to the conditions (d) given in Definition 1:

\begin{equation}
    h_{\mathcal{N}(v)}^{0}=h_{\mathcal{N}(v^{'})}^{0},\quad
    h_{v}^{1}=h_{v^{'}}^{1},
\end{equation}

According to the above equations and the mathematical inductive. the representations of $v\in V, v^{'}\in V^{'}$ of the $l$-th layer meet the condition:

\begin{equation}
    h_{v}^{l}=h_{v^{'}}^{l},
\end{equation}

For a GNN with $K$ layers, the representations of $G$ and $G^{'}$ are identical:

\begin{equation}
h_G=READ(\{ h^k_v|v\in G \} ), h_{G^{'}}=READ(\{ h^k_{v^{'}}|v^{'}\in G^{'} \} ).
\end{equation}

\begin{equation}
h_{G}=h_{G^{'}}.
\end{equation}

Obviously, existing GNN provides identical representation for a pair of PNG: $h_{G}=h_{G^{'}}$. The reason that GNNs are confused by PNGs is due to their adopted neighbor aggregation strategy, which captures node-level information only. Unfortunately, PNGs share identical characteristics at the node-level but have different global structures. As a result, GNNs can not distinguish PNGs effectively.

{\color{red}{Garg et al.\cite{garg2020generalization} have indicated that the LPI problem limits the representational power of GNNs. Theorem 2 further indicates how the LPI problem affects the graph classification performance of GNNs from the perspective of graph isomorphism. Another interesting question is how the LPI problem influences the node classification task. Unlike the graph structure which is implicated in the LPI problem, the node classification task is unaffected by global structural information from the whole graph. Therefore, to what extent and precisely how the LPI problem influences node classification needs deeper mathematical analysis and associated proofs, which are beyond the scope of this paper we will we hope to investigate in more detail in further work. In this paper, we simply focus on the LPI problem for the graph classification task.}}

However, although PNGs have identical node-level characteristics, their subgraph-level characteristics are different. The following lemmas demonstrate that the structural differences between PNGs can be learned from their generated graphs. In other words, the subgraph-level information is helpful in distinguishing PNGs.

\textbf{Lemma 3. }
For two permutation non-isomorphic graphs $G=(V,E)$ and $G^{'}=(V^{'},E^{'})$, their generated graphs $Gg=(Vg,Eg)$ and $Gg^{'}=(Vg^{'},Eg^{'})$ are structurally distinct.


\begin{figure*}
    \centering
    \includegraphics[scale=0.45]{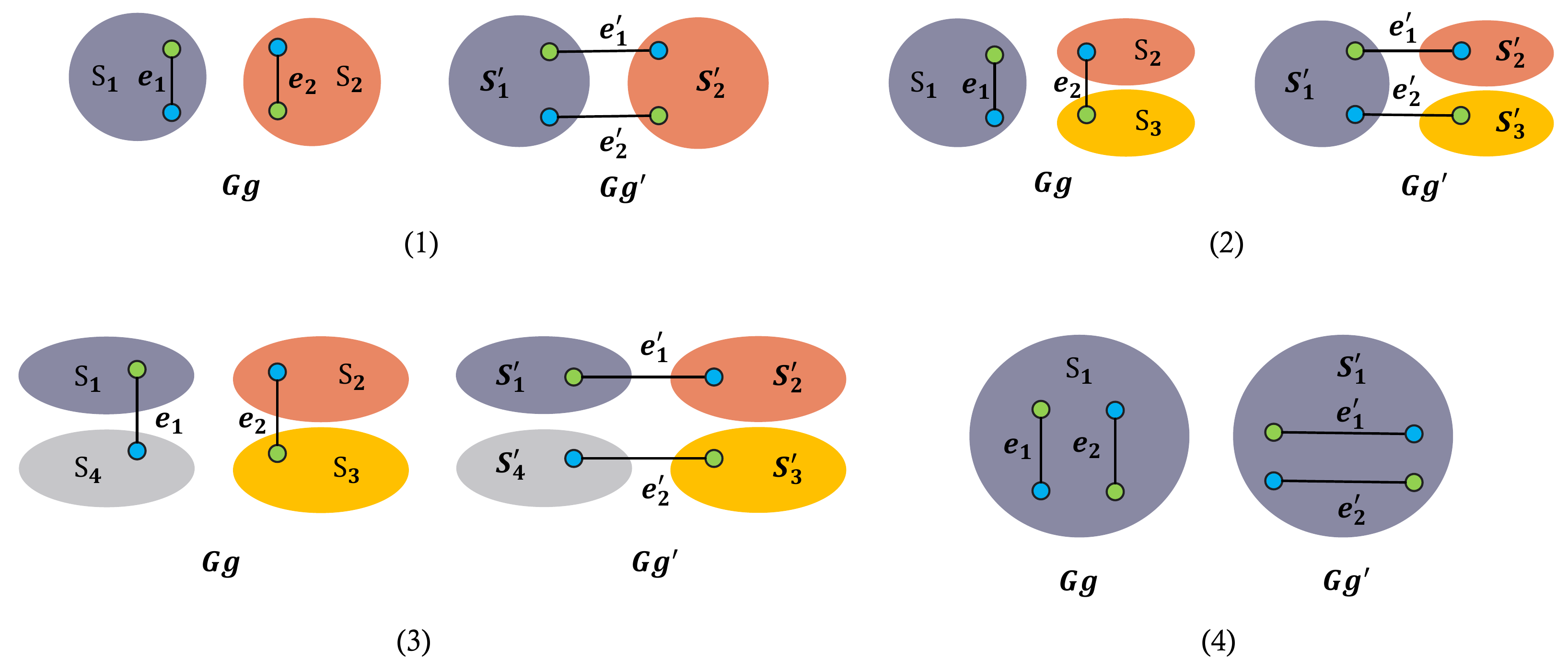}
    \caption{Four relations between edges and subgraphs. $S_i$ is the subgraph with the index $i$. Blue and green points are nodes. The same color nodes share the same feature.}
    \label{Lemmma3 relations}
\end{figure*}

We provide a proof of Lemma 3. by analyzing the relationship between edges and subgraphs. For the edges:

\begin{equation}
e_i\in E, \quad e_{i}^{'}\in E^{'}, \quad (i=1,2),
\end{equation}

\begin{equation}
e_1=(a,b), \quad e_2=(i,j), \quad e_{1}^{'}=(a,i), \quad e_{2}^{'}=(b,j),
\end{equation}

there are four types of relations between them.

\textbf{Relation 1:}

As shown in Fig.~\ref{Lemmma3 relations} (1), for the subgraphs $S_i$ and their nodes $V_i$ and edges $E_i$:

\begin{equation}
S_i=(V_i,E_i), \quad S_{i}^{'}=(V_{i}^{'}, E_{i}^{'}), \quad (i=1,2).
\end{equation}

Relation 1 can be written as:

\begin{equation}
e_i\in E_i, e_{i}^{'}\notin E_{i}^{'}, i=1,2.
\end{equation}

Due to the features of supernodes (subgraphs) include node counts and the subgraph type respectively (Section.~\ref{Generating Graphs}), the edges change must lead to node counts or subgraph change of subgraphs. We thus have:

\begin{equation}
E_i\ne E_{i}^{'}, h_{S_{1}}^{0}\ne h_{S_{1}^{'}}^{0},
\end{equation}

and so, $Vg\ne Vg^{'}$, $Gg$ and $Gg^{'}$ are different.

\textbf{Relation 2:}

As shown in Fig.~\ref{Lemmma3 relations} (2), consider the subgraphs:

\begin{equation}
S_i=(V_i,E_i), S_{i}^{'}=(V_{i}^{'},E_{i}^{'}), (i=1,2,3).
\end{equation}

Relation 2 can be written as:

\begin{equation}
e_1\in E_1, e_{i}^{'}\notin E_{j}^{'}, e_{2}\notin E_{j},
\end{equation}

\begin{equation}
i=1,2 \quad j=1,2,3.
\end{equation}

Due to the relation:

\begin{equation}
    e_1\in E_1, e_{1}^{'}\notin E_{1}^{'}.
\end{equation}

We have:

\begin{equation}
V_{i}=V_{i}^{'}, (i=1,2,3),
\end{equation}

\begin{equation}
E_1\ne E_{1}^{'}, E_{j}=E_{j}^{'}, (j=2,3). 
\end{equation}

So, $S_1\ne S_{1}^{'}$, $h_{S_{1}}^{0}\ne h_{S_{1}^{'}}^{0}$. $Gg$ and $Gg^{'}$ are different.

\textbf{Relation 3:}

As shown in Fig.~\ref{Lemmma3 relations} (3), for the subgraphs:

\begin{equation}
S_i=(V_i,E_i), S_{i}^{'}=(V_{i}^{'},E_{i}^{'}), (i=1,2,3,4).
\end{equation}

Relation 3 can be written as:

\begin{equation}
e_i\notin E_j, e_{i}^{'}\notin E_{j}^{'}, E_{j}=E_{j}^{'}.
\end{equation}

\begin{equation}
i=1,2 \quad j=1,2,3,4.
\end{equation}

$Gg=Gg^{'}$ if and only if:

\begin{equation}
    \left\{
        \begin{array}{cc}
             h_{S_{1}}^{0}=h_{S_{1}^{'}}^{0}=h_{S_{3}}^{0}=h_{S_{3}^{'}}^{0}&  \\
             h_{S_{2}}^{0}=h_{S_{2}^{'}}^{0}=h_{S_{4}}^{0}=h_{S_{4}^{'}}^{0}& \\
             \exists e_3=(S_{1},S_{4}), \exists e_4=(S_{2},S_{3}),\\                             \exists e_5=(S_{1},S_{2}), \exists e_6=(S_{3},S_{4})&
        \end{array}
    \right. 
\end{equation}

Obviously, $G=G^{'}$ when $Gg=Gg^{'}$. Because swapping the nodes constituting edges does not change the connection structure of the subgraphs and their nodes, the structures are identical.

Else, $Gg\ne Gg^{'}$, because of the super-node features or connections of super-nodes.

\textbf{Relation 4:}

As shown in Fig.~\ref{Lemmma3 relations} (4), for the subgraphs:

\begin{equation}
S_1=(V_1,E_1), S_{1}^{'}=(V_{1}^{'},E_{1}^{'}),\quad e_1\in S_1, e_{1}^{'}\in S_{1}^{'}.
\end{equation}

Obviously, $S_1, S_{1}^{'}\notin Tree$. If $S_{1}\in Path, S_{1}^{'}\in Cir^{'}$ or $S_{1}\in Cir, S_{1}^{'}\in Path^{'}$:

\begin{equation}
h_{S_{1}}^{0}\ne h_{S_{1}^{'}}^{0}.
\end{equation}

There is a special case of Lemma 3 which deserves comment. If $S_{1}$ is a Path or a Cir, changing a pair of edges would divide $S_{1}$ into two subgraphs or lead to a subgraph identical to $S_{1}$ but with a different node sequence. We will discuss this situation in Lemma 4.

\begin{figure}[ht]
    \centering
    \includegraphics[scale=0.45]{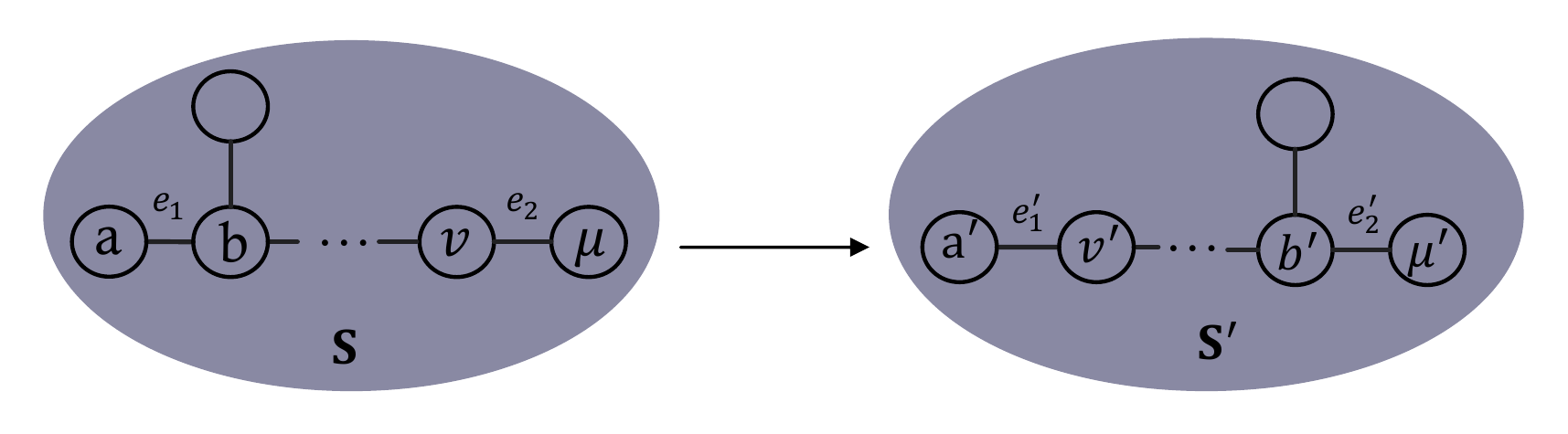}
    \caption{A special case. Changing a pair of edges, $e_{1}$ and $e_{2}$, leads to different node sequences. Subgraphs $S$ and $S^{'}$ have identical structure but different node sequences. }
    \label{Lemmma4 case}
\end{figure}

\textbf{Lemma 4. }
GNNs can distinguish the structural variance caused by the node sequence.

In summary, because GNN adopts a hierarchical aggregation strategy, different node sequences will lead to different information aggregated by GNN in different layers.

As shown in Fig.~\ref{Lemmma4 case}. If $u,v$ are $i$-hop and $j$-hop neighbors of node $a$ respectively, then $u^{'},v^{'}$ are respectively $j$-hop and $i$-hop neighbors of node $a^{'}$. Then as a result $h_{a}^{i}\ne h_{a^{'}}^{i}$ and $h_{a}^{j}\ne h_{a^{'}}^{j}$. Therefore, GNNs can distinguish them.

\textbf{Lemma 5. }As for Lemma 3, if $Gg=(Vg,Eg)$ and $Gg^{'}=(Vg^{'},Eg^{'})$ are different in terms of their supernode features or subgraph connections, then a GNN can distinguish the generated graphs.

Lemma 3 and Lemma4 lead to Lemma 5. Due to the properties of GNNs, Lemma 5 can be demonstrated easily. 

Suppose $S^{'}$ is the permuted subgraph of $S\in V_g$. Under relations 1,2 and 4, $\exists S\in Vg, S^{'}\in Vg^{'}$ and let the equation below be tenable:

\begin{equation}
    h_{S}^0\neq h_{S^{'}}^0.
\end{equation}

Similar to the proof of Theorem.~2, we can demonstrate $h_{G_g}\neq h_{G_g'}$ via the mathematical inductive.

For relation 3, we have:

\begin{equation}
\mathcal{N}(S)\neq \mathcal{N}(S^{'}),
\end{equation}

which means:

\begin{equation}
h_{\mathcal{N}(S)}\neq h_{\mathcal{N}(S^{'})},
\end{equation}

After combining operations, we have:

\begin{equation}
h_{S}^1= COM(h_{S}^0,h_{N(S)}), \quad h_{S^{'}}^1= COM(h_{S^{'}}^0,h_{N(S^{'})}).
\end{equation}

Due to $COM$ is an injective function, different inputs will be mapped into different points in the feature space. It means that graph representations after iterations are different:

\begin{equation}
h_{S}^1\neq h_{S^{'}}^1,\quad h_{G_g}\neq h_{G_g'}.
\label{diff graph rep}
\end{equation}

Eq.~\eqref{diff graph rep} shows that PNGs can be distinguished by the differences between subgraphs, which proves the Lemma 5. According to the above theoretical analysis, we propose the framework of TL-GNN and discuss it in the next subsection.

\subsection{The TL-GNN Framework}
\label{Framework of TL-GNN}
As noted above the LPI of GNNs lead to serious representational limitations. Existing GNNs are therefore sometimes compromised in their performance by PNGs. Fortunately and as we have shown, the subgraph-level information can be beneficial in distinguishing the PNGs. To overcome the limitations caused by the LPI problem and enrich the representational capacity of GNNs, we propose a novel GNN approach which we refer to as the Two-level Graph Neural Network(TL-GNN). In this section, we present details of the TL-GNN framework.

\begin{figure*}
    \centering
    \includegraphics[scale=0.5]{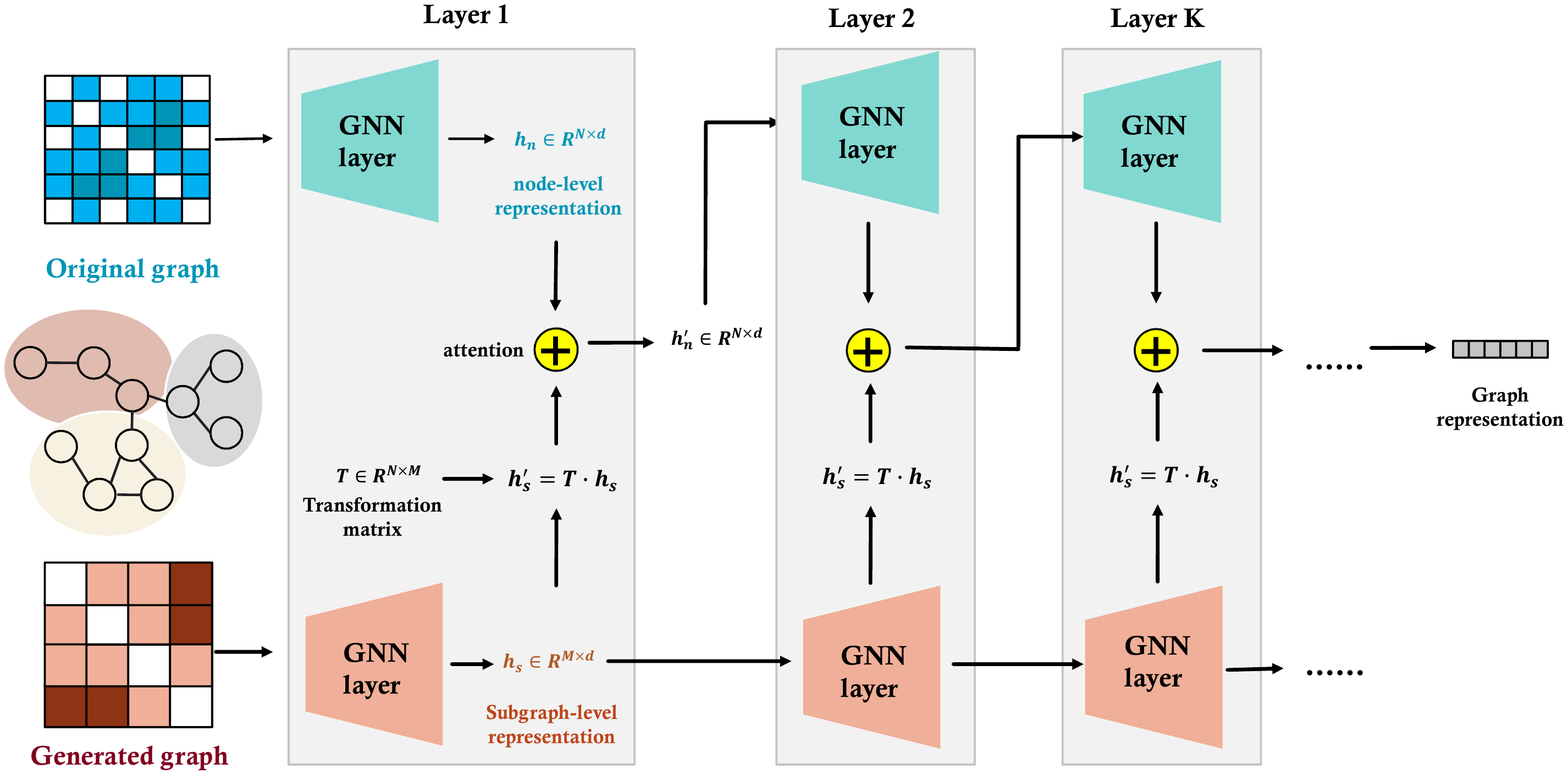}
    \caption{Visual illustration of the TL-GNN framework. The model takes both the original graph and generated graph as input, then aggregates them with the same strategy. Each TL-GNN layer has two separate GNN blocks for graph and generated graph to capture node-level and subgraph-level information respectively. At the end of each layer, the subgraph-level representation $h_s$ would be converted into $h_s'$ via the transformation matrix $T$, which could be aligned with node-level representation $h_n$. On this basis, two independent representations can be merged together with the attention mechanism. The summation $h_n'$ would be the input of the next layer.}
    \label{Framework}
\end{figure*}

Due to the neighbor aggregation strategy, existing GNNs focus on capturing node-level information and ignore higher level structural arrangement information. Unlike existing GNNs, the TL-GNN can capture both node-level information and subgraph-level arrangement information simultaneously. This is achieved by merging subgraph-level information into supernode-level information.

Specifically, the TL-GNN accepts both a graph and its generated graph (as described in Section~\ref{Constructing Generated Graph}) as inputs. In each layer, the two separate GNN components capture information concerning the graph (node-level) and generated graph (subgraph-level) respectively, as illustrated in Fig.\ref{Framework}.

The node-level propagation process can be described as:

\begin{equation}
    \begin{cases}
    h_{\mathcal{N}(v)}^{k}=AGG(\{h_\mu^{(k-1)},\forall\mu\in\mathcal{N}(v)\}),\\
    \\
    \widetilde{h_{v}^{k}}=COM(h_v^{k-1},h_{\mathcal{N}(v)}^{k}),\\
    \end{cases}
\end{equation}

On the other hand, the subgraph-level propagation can be described as:

\begin{equation}
    \begin{cases}
    h_{\mathcal{N}(s)}^{k}=AGG(\{h_\gamma^{(k-1)},\forall\gamma\in\mathcal{N}(s)\}),\\
    \\
    h_{s}^{k}=COM(h_s^{k-1},h_{\mathcal{N}(s)}^{k}),\\
    \end{cases}
\end{equation}

Here $\mathcal{N}(v)$ and $\mathcal{N}(s)$ are the neighbor sets of node $v$ and supernode $s$ respectively. These two information propagation processes are discrete and do not share parameters.

\subsubsection{The AGG\_SUB and MERG functions}
With a two-level representation to hand, we merge the subgraph-level representation into the node-level representation. We design two functions AGG\_SUB and MERG for this process.

For a graph containing $N$ nodes, assume that its generated graph contains $M$ supernodes. The outputs of the $k-$th layer of TL-GNN are the node-level representation $\widetilde{h}_{N}^{k}$ and the subgraph-level representations $h_{S}^{k}$ respectively:

\begin{equation}
\widetilde{h}_{N}^{k} \in \mathbb{R}^{N\times d}, \quad h_{S}^{k} \in \mathbb{R}^{M\times d}, 
\end{equation}

Here $\widetilde{h}_{N}^{k}$ and $h_{S}^{k}$ are matrices that contain the node and supernode (subgraph-level) representations respectively. Each row of $h_{N}^{k}$ or $h_{S}^{k}$ corresponds to the representation of an individual node or supernode. 

We merge (or concatenate) the matrices $h_{N}^{k}$ and $h_{S}^{k}$ into a single matrix $\widetilde{h_{N}^{k}}$ using the MERG function. Then $\widetilde{h_{N}^{k}}$ and $h_{S}^{k}$ become the inputs into the next layer of the GNN. Each layer has an identical matrix structure. After applying the merging step, the representations of the nodes are merged with the available subgraph-level information.

Next, we define two more functions for the TL-GNN, namely AGG\_SUB and MERG:

\begin{equation}
    H_{s}^{k}=AGG\_SUB(h_{S}^{k})\label{AGG_SUB},\quad
    h_{N}^{k}=MERG(\widetilde{h_{N}^{k}}, H_{S}^{k}),
\end{equation}

The AGG\_SUB operator aggregates all of the subgraph representations which contain a given node $v$. The output of the MERG function is the matrix $h_{N}^{k}$, which merges the node-level and subgraph-level information into a single representation.

Finally, the representation of the graph $G$ is obtained from the representations of its nodes. 

\begin{equation}
    h_G=READ(\{ \widetilde{h^k_v}|v\in G \} ),
\end{equation}

In order to distinguish PNGs, the AGG\_SUB and MERG need to fulfill conditions specified in Lemma 6:

\textbf{Lemma 6. }If $AGG\_SUB$ and $MERG$ are both injective multiset functions, then the representation of $G$ is distinct from that of $G^{'}$.

To demonstrate Lemma 6, We define $C(v)$, which is a set of subgraphs that contain node $v$. According to Lemma 5:


\begin{equation}
\exists S\in C(v), S^{'}\in C(v'), \quad h_{S}^1\neq h_{S^{'}}^1,
\end{equation}

As 
\begin{equation}
    h_v^0=h_{v'}^0,\quad h_{c(v)}^0\neq h_{c({v^{'}})}^0,
\end{equation}

\begin{equation}
    h_v^1=MERG(h_v^0, h_{c(v)}^0),\quad h_{v^{'}}^1=MERG(h_{v^{'}}^0, h_{c({v^{'}})}^0),
\end{equation}

\begin{equation}
    h_v^1\neq h_{v'}^1,
\end{equation}

The above equations are satisfied if and only if $MERG$ is an injective function. When the above equations are satisfied, we can obtain $h_G\neq h_{G'}$, which means $G$ and $G'$ can be distinguished. Lemma 6 is proved.

There are several choices available for the AGG\_SUB and MERG functions, which include summation and concatenation. We chose summation for AGG\_SUB of TL-GNN. 

Due to the fact that the graph size and generated graph size are different, the subgraph-level representations need to be transformed into node-level representations of identical size.

The translation matrix mentioned in Section~\ref{Constructing Generated Graph} translates the matrix $h_{S}^{k}$ into the matrix $H_{S}^{k} \in \mathbb{R}^{N\times d}$:

\begin{equation}
H_{S}^{k}=T\cdot h_{S}^{k},
\end{equation}

Due to the definition of $T$, the $i$-th row of $H_{S}^{k}$ is the summation of representations of those supernodes that  contain node $i$. Assume $i$-th line of $H_{S}^{k}$ is $(H_{s}^{k})_{i}$:

\begin{equation}
    (H_{s}^{k})_{i}=\sum_{s\in\mathcal{C}(i)}h_s^{k},
\end{equation}

where $\mathcal{C}(i)$ is the set of supernodes (subgraphs) that contain node $i$.

As for the MERG operation, we use an attention mechanism to define the MERG function. The attention mechanism for TL-GNN is essentially a weighted summation. Therefore,the MERG operation can be rewritten as:

\begin{equation}
    h_{v}^{k}=\alpha^k \cdot \widetilde{h_{v}^{k}}+\beta^k \cdot H_{s}^{k},
\end{equation}

\begin{equation}
    \alpha^k=\frac{exp(\hat{\alpha}^k)}{exp(\hat{\alpha}^k)+exp(\hat{\beta}^k)},\quad \beta^k=\frac{exp(\hat{\beta}^k)}{exp(\hat{\alpha}^k)+exp(\hat{\beta}^k)},
\end{equation}

where $\hat{\alpha}^k$ and $\hat{\beta}^k$ are randomly initialized scales. $\alpha$ and $\beta$ satisfy the condition $\alpha^k+\beta^k=1$. The parameters $\alpha^k$ and $\beta^k$ can be learned during training. The parameter $\alpha^k$ is large if the node-level representation is more important for the classification of graphs, and vice versa.

Finally, the effectiveness of TL-GNN in distinguishing PNGs can be stated mathematically.

\textbf{Theorem 7.}
TL-GNN has the ability to distinguish the PNG. 

We provide the proof of Theorem 7 using the above Lemmas. 

For permutation non-isomorphic graphs $G=(V,E)$ and $G^{'}=(V^{'},E^{'})$, According to Lemma 6, we have:

\begin{equation}
h_v^1\neq h_{v'}^1 \ v\in\ V\ v^{'}\in\ V^{'}.
\end{equation}

According to mathematical induction, the representations of $v\in V, v^{'}\in V^{'}$ of $l$-th layer meet condition:

\begin{equation}
    h_{v}^{l}=h_{v^{'}}^{l}.
\end{equation}

For a GNN with $K$ layers, the representations of $G$ and $G^{'}$ are different:

\begin{equation}
h_G=READ(\{ h^k_v|v\in G \} ),\quad h_{G^{'}}=READ(\{ h^k_{v^{'}}|v^{'}\in G^{'} \} ),
\end{equation}

\begin{equation}
h_{G}\neq h_{G^{'}}.
\end{equation}

So, TL-GNN can distinguish permutation non-isomorphic graphs $G$ and $G^{'}$.

Theorem 7 indicates that the TL-GNN can distinguish those ambiguous graphs that confuse existing GNNs. In other words, TL-GNN is more powerful than GNNs.

\section{Experiments}
\label{Experiments}
In this section, we perform experimental evaluations of our TL-GNN method on the graph classification task. We compare the TL-GNN to several state-of-the-art deep learning and graph kernel methods, and conduct experiments on seven standard graph classification benchmarks together with synthetic data.

\subsection{Datasets}
Datasets of this paper include MUTAG\cite{debnath1991structure}, PTC\cite{toivonen2003statistical}, NCI1\cite{9206723}, PROTEINS\cite{R1990Protein}, COX2\cite{sutherland2003spline}, IMDB\_ M\cite{DBLP:journals/corr/abs-1904-12189} and IMDB\_B\cite{cai2018simple}. The IMDB\_ M and IMDB\_ B datasets have no node features. The remaining datasets have categorical node features. In order to verify the ability to distinguish PNGs, we have prepared a synthetic PNG dataset named SPNG. The details of these datasets are shown in Appendix.

\subsection{Baselines for Comparison}
The baselines used for comparison include state-of-the-art methods which are applied to the graph classification task:

(1) The kernel based methods: Weisfeiler-Lehman(WL) \cite{shervashidze2011weisfeiler} and subgraph Matching Kernel (CSM) \cite{articleCSM}, Deep Graph Kernel (DGK)\cite{yanardag2015deep}.

(2) The state-of-the-art GNNs: Graph convolution network (GCN) \cite{kipf2016semi}, Deep Graph CNN (DGCNN) \cite{ying2018graph}, Graph Isomorphism Network (GIN) \cite{xu2018powerful}, Random Walk Graph Neural Network (RW-GNN) \cite{nikolentzos2020random}, Graph Attention Network (GAT) \cite{velickovic2018graph}, Motif based Attentional Graph Convolutional Neural Network (MA-GCNN) \cite{peng2020motif}.

\begin{table*}[ht]\footnotesize
\centering
\large
\caption{\centering{CLASSIFICATION ACCURACY (IN \% $\pm$ STANDARD ERROR)}}
\resizebox{\textwidth}{!}{
\begin{tabular}{cccccccc}
\\\hline

Datasets &MUTAG &PTC &NCI1 &IMDB-M &IMDB-B &COX2 &PROTEINS\\ \hline

WL& $90.4\pm5.7$& $59.9\pm4.3$& \bm{$86\pm1.8$}& $50.9\pm3.8$& $73.8\pm3.9$& $83.2\pm0.2$& $75.0\pm3.1$ \\

CSM& $85.4$& $63.8$& $65.5$& $63.3$& $58.1$& $80.7\pm0.3$& -\\

DGK& $87.4\pm2.7$& $60.1\pm2.6$& $80.3\pm0.5$& $43.9\pm0.4$& $65.9\pm1.0$& -& $71.7\pm0.6$\\

GCN& $85.6\pm5.8$& $64.2\pm4.3$& $80.2\pm2.0$& $51.9\pm3.8$& $74.0\pm3.4$& -& $76.0\pm{3.2}$\\

DGCNN& $85.8\pm1.7$& $58.6\pm2.5$& $74.4\pm0.5$& $47.8\pm0.9$& $70.0\pm0.9$& -& $70.9\pm2.8$\\

GIN& $89.4\pm5.6$& $64.6\pm7.0$& $82.7\pm1.7$& $52.3\pm2.8$& $75.1\pm5.1$&$83.3\pm5.3$&$76.2\pm2.8$\\

FDGNN& $88.5\pm3.8$& $63.4\pm5.4$& $77.8\pm2.6$& $50.0\pm1.3$& $72.4\pm3.6$& $83.4\pm2.9$& $76.8\pm 2.9$\\

RW-GNN& $89.2\pm4.3$& $61.6\pm9.5$& $-$& $47.8\pm3.8$& $70.8\pm4.8$& $81.6\pm4.7$& $74.7\pm3.3$\\

GAT& $89.4\pm6.1$& $66.7\pm5.1$& $75.2\pm3.3$& $47.8\pm3.1$& $70.5\pm2.3$& -& $74.7\pm2.2$\\ 

HA-GCNN& $93.9\pm5.2$& $71.8\pm6.3$& $81.8\pm2.4$& $53.8\pm3.1$& $77.2\pm3.0$& -& $79.4\pm1.7$\\ \hline

TL-GNN\_sm& $90.9\pm6.4$& $68.1\pm5.0$& $81.9\pm3.3$& $54.3\pm4.7$& $77.5\pm2.0$& $86.7\pm3.5$& $77.3\pm2.6$\\ 

TL-GNN\_ms& $91.2\pm3.9$& $67.0\pm7.9$& $82.1\pm4.2$& $53.4\pm4.7$& $77.5\pm3.5$& $86.2\pm4.6$& $78.9\pm2.3$\\ 

TL-GNN\_mm& $90.8\pm5.4$& $66.3\pm7.6$& $81.0\pm3.6$& $52.2\pm3.8$& $76.6\pm3.3$& $85.6\pm2.3$& $77.5\pm2.6$\\ 

TL-GNN(w/o S)& $92.4\pm6.3$& $70.0\pm7.9$& $82.2\pm4.9$& $54.4\pm3.0$& $77.8\pm2.1$& $87.8\pm2.7$& $79.4\pm3.0$\\ 

TL-GNN& \bm{$95.7\pm3.4$}& \bm{$74.4\pm4.8$}& $83.0\pm2.1$& \bm{$55.1\pm3.2$}& \bm{$79.7\pm1.9$}& \bm{$88.6\pm2.7$}& $\bm{79.9\pm4.4}$\\ 

\hline
\end{tabular}}
\label{test acc}
\end{table*}

\subsection{Experimental Setup}
For our experimental comparison, we set the GNN layers of GIN so as to have the same structure but with no parameter sharing. These layers aggregate and combine the original graph and its generated graph. There are independent attention parameter pairs for merging operations between the GNN layers. Each GNN layer has several MLP layers. More details about experimental setup are shown in Appendix. 

\subsection{Results and Discussion}

\textbf{Comparison with existing GNNs on real-world datasets:} 

The results in Table \ref{test acc} indicate that TL-GNN achieves the best results on 6 out of 7 benchmarks, often with a clear improvement over alternative GNN methods studied. The performances for the classical GNNs are quoted from their indicated reference. We perform 10-fold cross-validation to compute the GIN, GCN and RW-GNN accuracies on COX2. The parameters for the deep learning methods are as suggested by their authors. For fairness, all the methods run on the same computing device. For cases where accuracy cannot be obtained, we use the "-" tag in Table \ref{test acc}.

We found that TL-GNN always achieves the best results on datasets containing sparse graphs. For PTC and COX2, TL-GNN achieves $2.6\%$ and $5.2\%$ margins of improvements over the second-best method.  The accuracies of TL-GNN on MUTAG and IMDB-M are $95.7\%$ and $55.1\%$ respectively. This represents a slight but consistent improvement over the alternative methods studied. TL-GNN also achieves the best performance on PROTEINS. Although TL-GNN gives only a slight improvement compared to the second-best method. The average degree of PROTEINS is more than 3. which is dense compared with the remaining datasets. For this kind of dense graph, TL-GNN can capture a large number of subgraphs and enrich the learned information. Even in the cases where TL-GNN does not achieve the best performance, its accuracy is close to that of the best performing method. For NCI1, TL-GNN achieves $0.3$\% more accuracy than the second best deep learning method. 

GIN and TL-GNN have identical GNN layers. However, TL-GNN achieves better performances on most of the datasets. For example, TL-GNN achieves $6.3\%$ and $9.8\%$ improvements on MUTAG and PTC compared to GIN. Moreover, the standard errors for TL-GNN on MUTAG and PTC are lower than those for GIN. NCI1 is the only dataset on which TL-GNN can not surpass the performance of WL. It is worth noting that all deep learning methods are also do not outperform WL on NCI1.

It is worth noting that both GAT and TL-GNN apply the attention mechanism. The difference is that GAT assigns attention weights to the neighbors of nodes. This means that GAT pays different attention to node-level information. Unlike GAT, TL-GNN assigns attention weights to different levels of information (both node-level and subgraph-level). The results show that TL-GNN outperforms GAT on all datasets, and it also achieves significant improvements on several datasets. For example, on NCI and IMDB, TL-GNN achieves $6-9\%$ improvement compared to GAT.

\begin{figure*}[htbp]
\centering
\subfigure[MUTAG.]{
\begin{minipage}[H]{0.2\linewidth}
\centering
\includegraphics[width=1.7in]{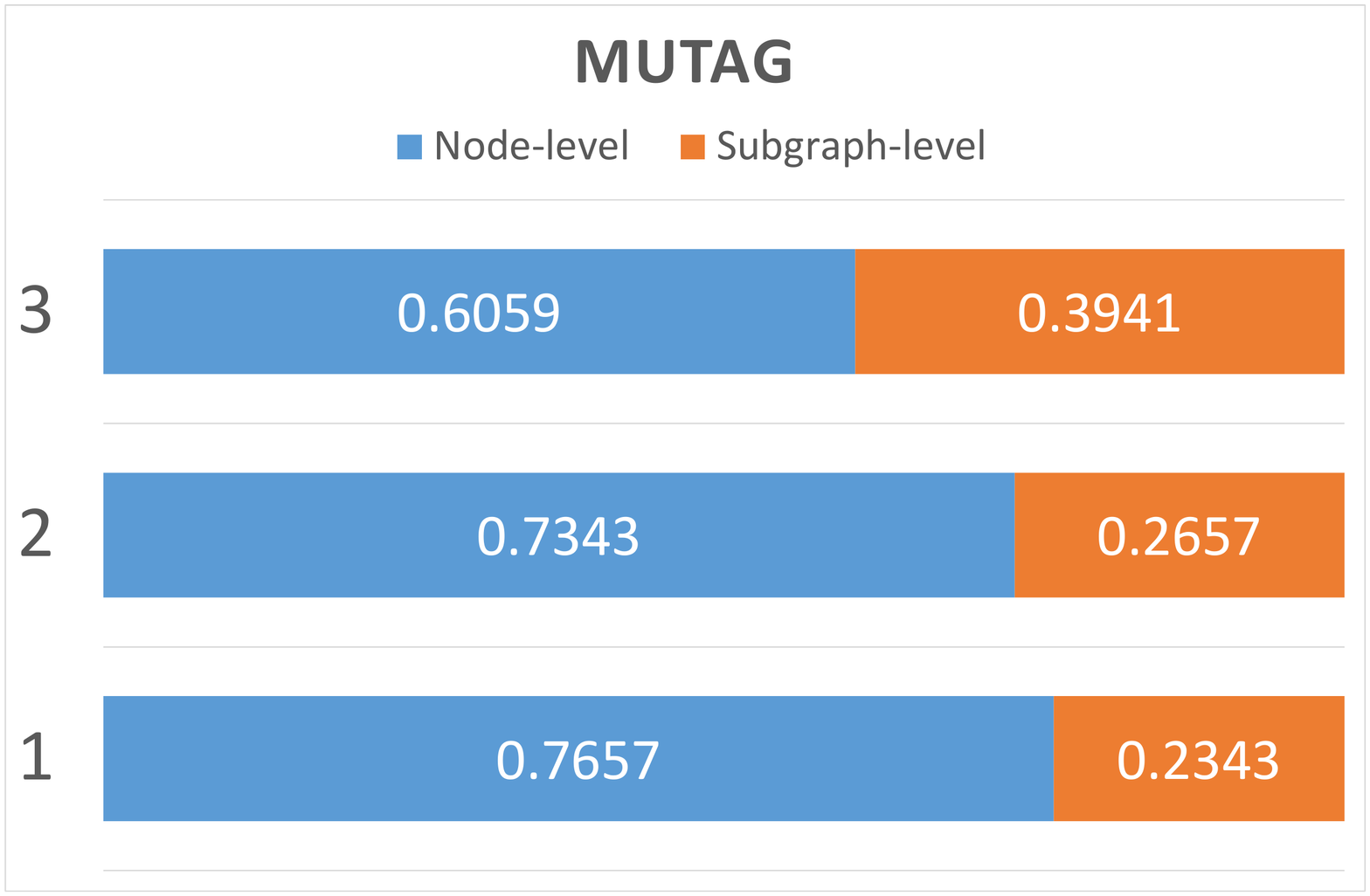}
\end{minipage}
}
\subfigure[PTC.]{
\begin{minipage}[H]{0.2\linewidth}
\centering
\includegraphics[width=1.7in]{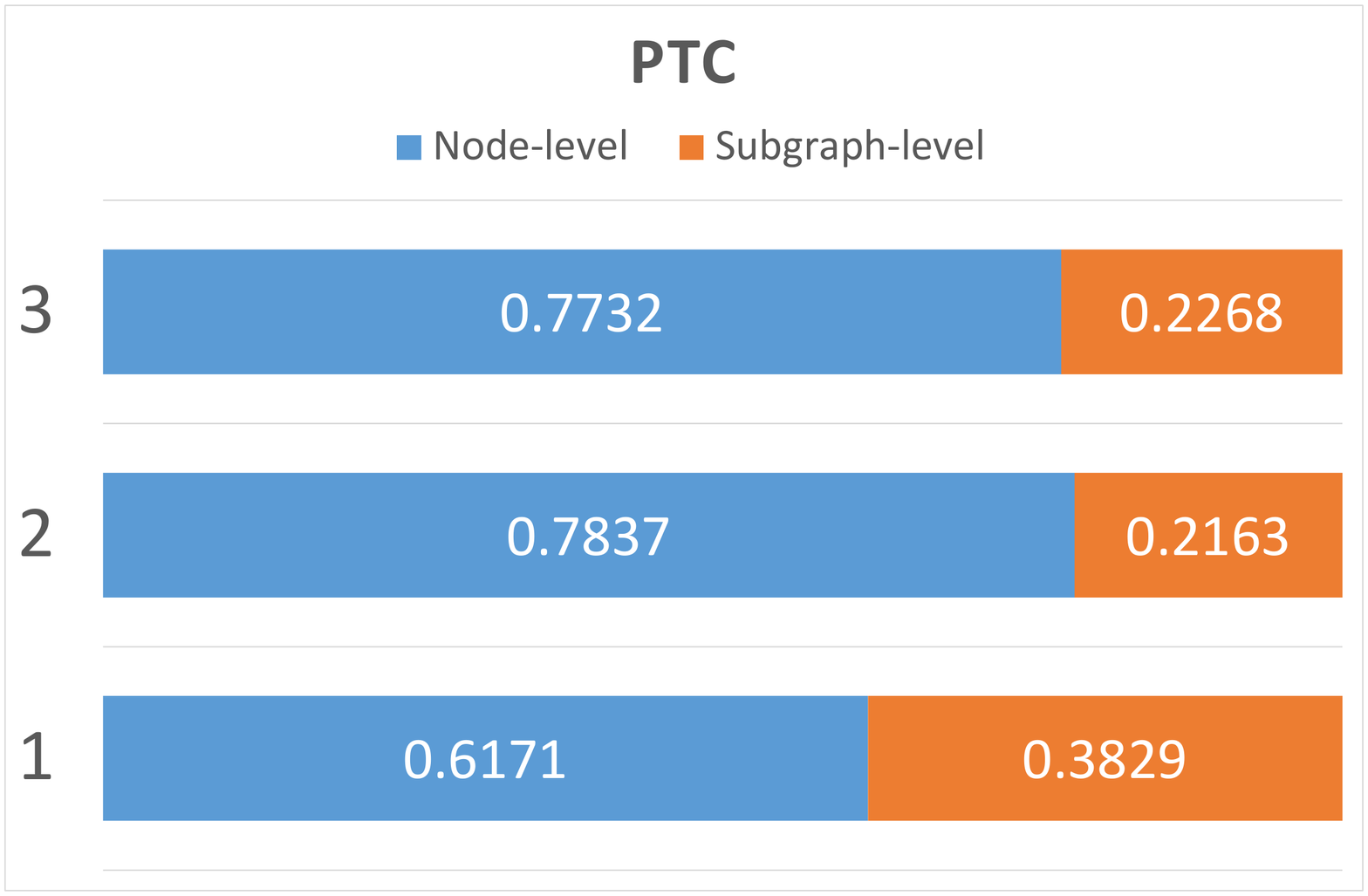}
\end{minipage}
}
\subfigure[COX2.]{
\begin{minipage}[H]{0.2\linewidth}
\centering
\includegraphics[width=1.7in]{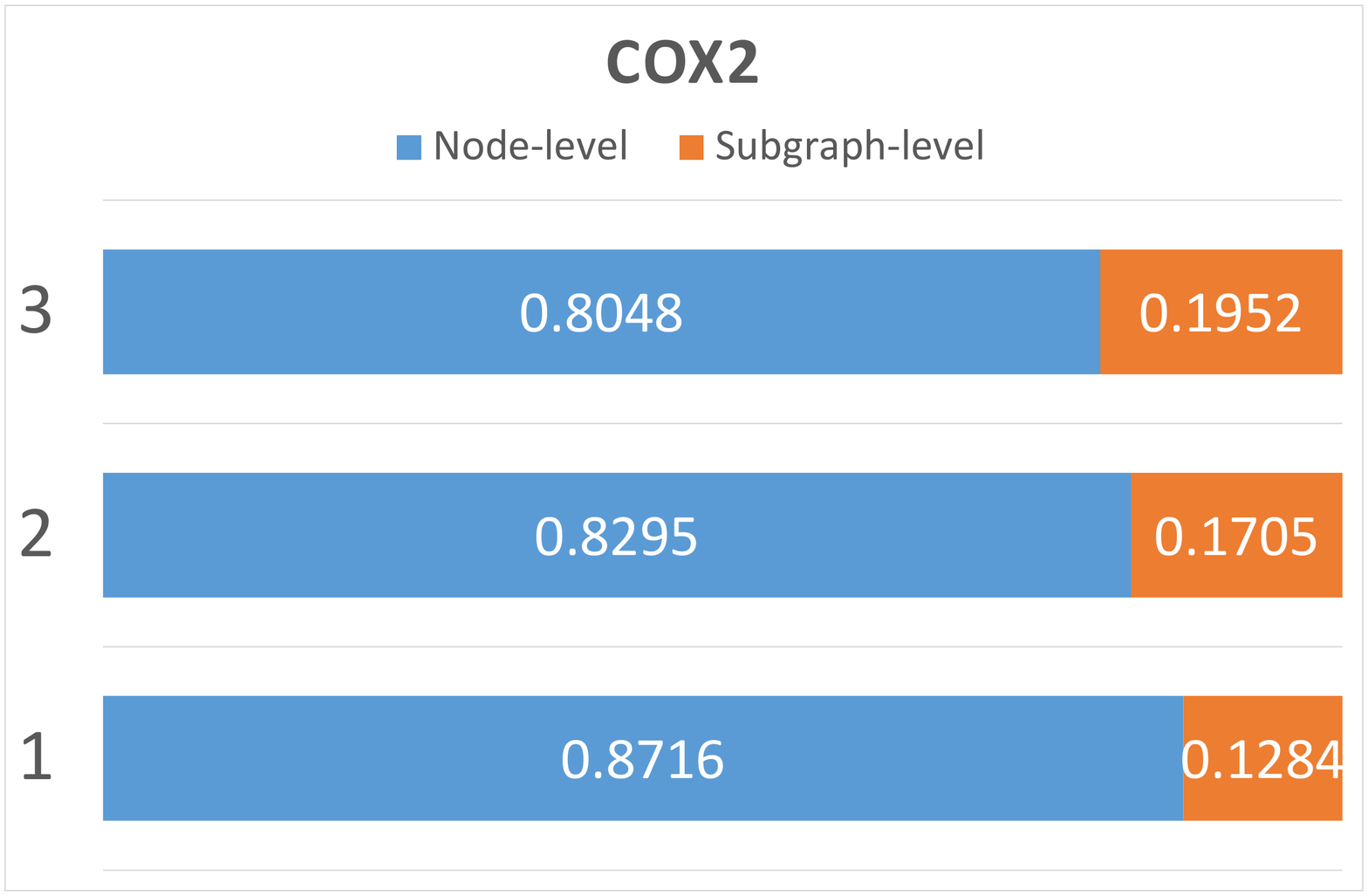}
\end{minipage}
}
\subfigure[SPNG.]{
\begin{minipage}[H]{0.2\linewidth}
\centering
\includegraphics[width=1.7in]{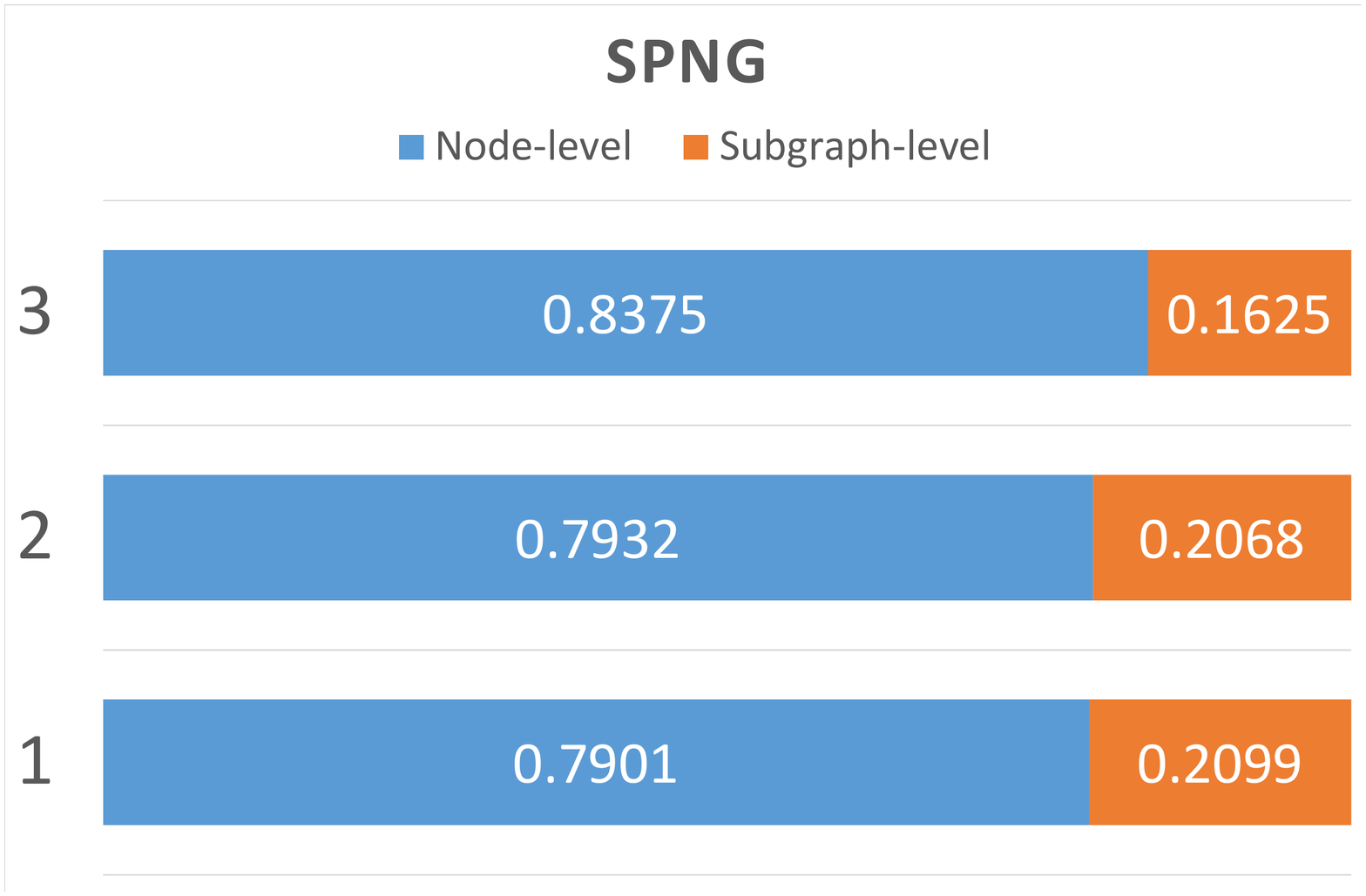}
\end{minipage}
}
\subfigure[NCI1.]{
\begin{minipage}[H]{0.2\linewidth}
\centering
\includegraphics[width=1.7in]{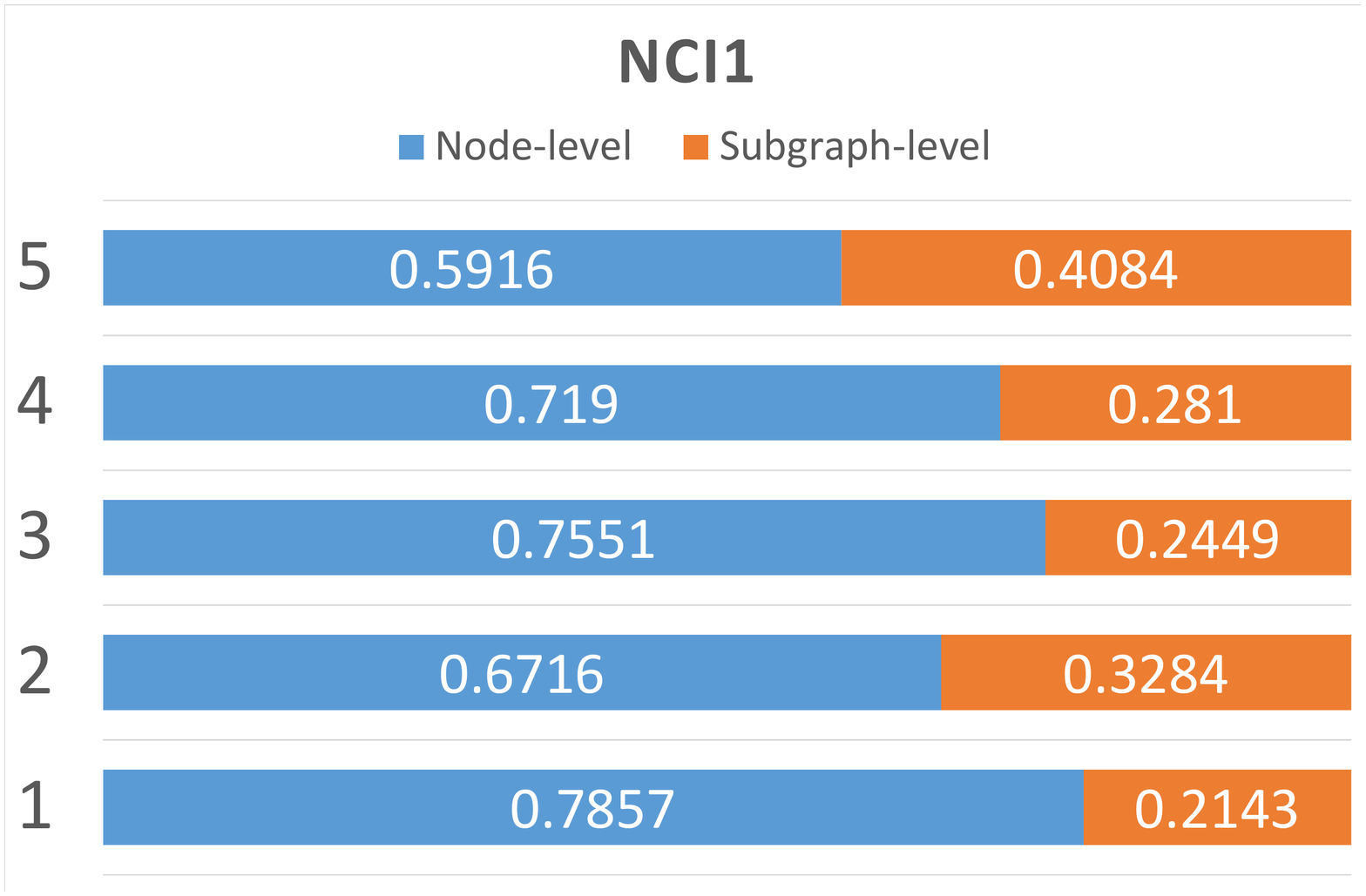}
\end{minipage}
}
\subfigure[PROTEINS.]{
\begin{minipage}[H]{0.2\linewidth}
\centering
\includegraphics[width=1.7in]{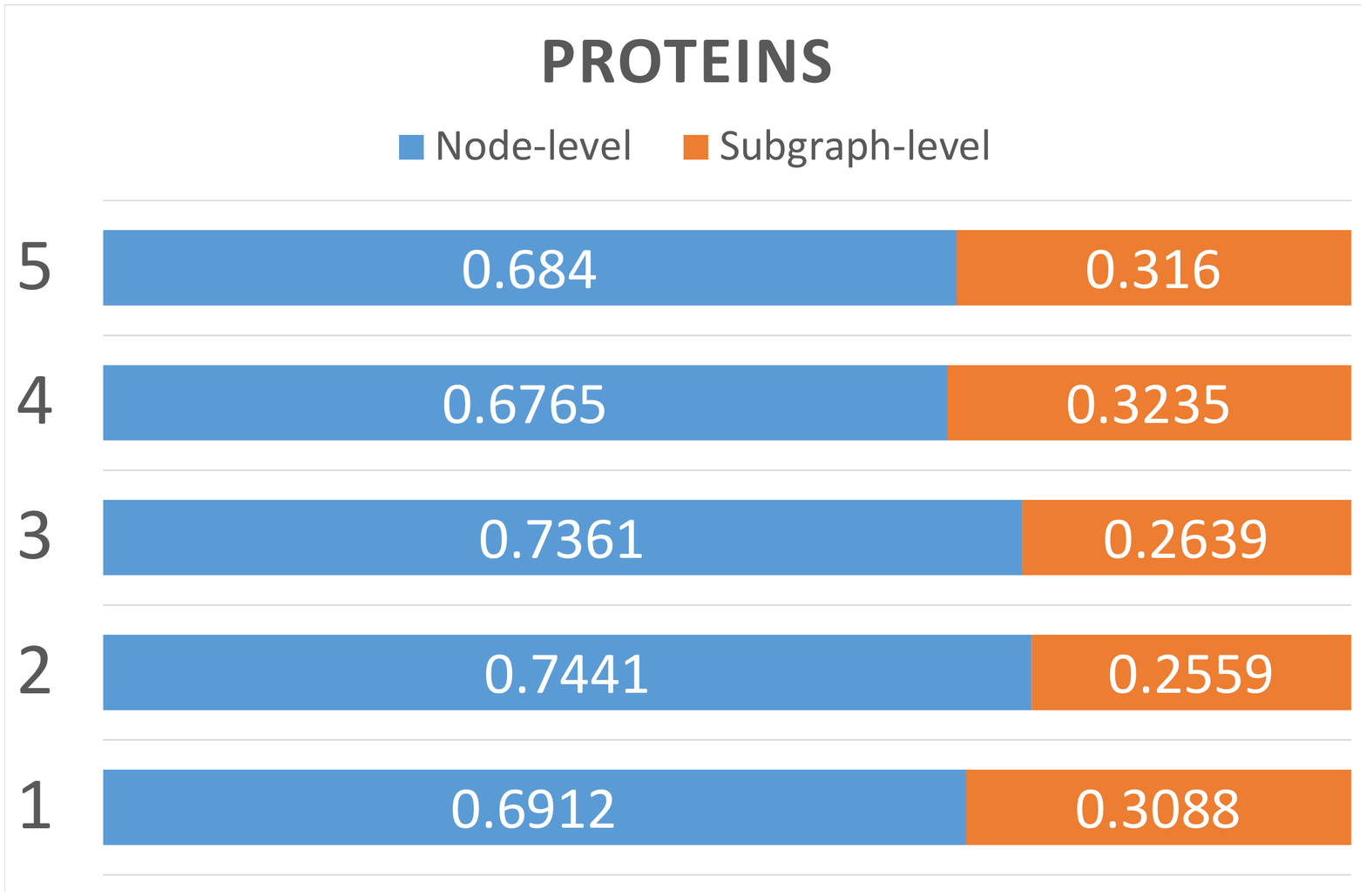}
\end{minipage}
}
\subfigure[IMDB-M.]{
\begin{minipage}[H]{0.2\linewidth}
\centering
\includegraphics[width=1.7in]{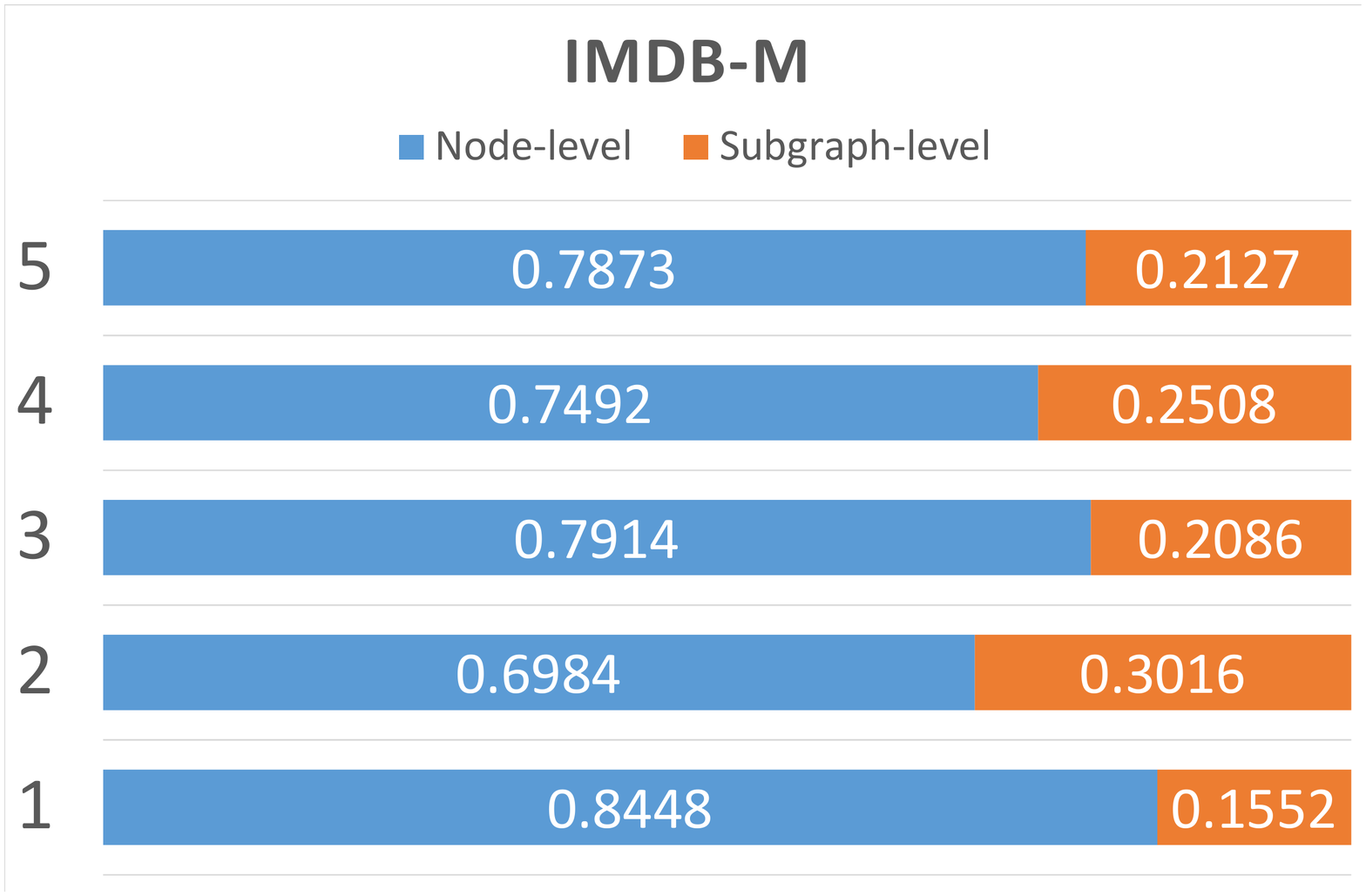}
\end{minipage}
}
\subfigure[IMDB-B.]{
\begin{minipage}[H]{0.2\linewidth}
\centering
\includegraphics[width=1.7in]{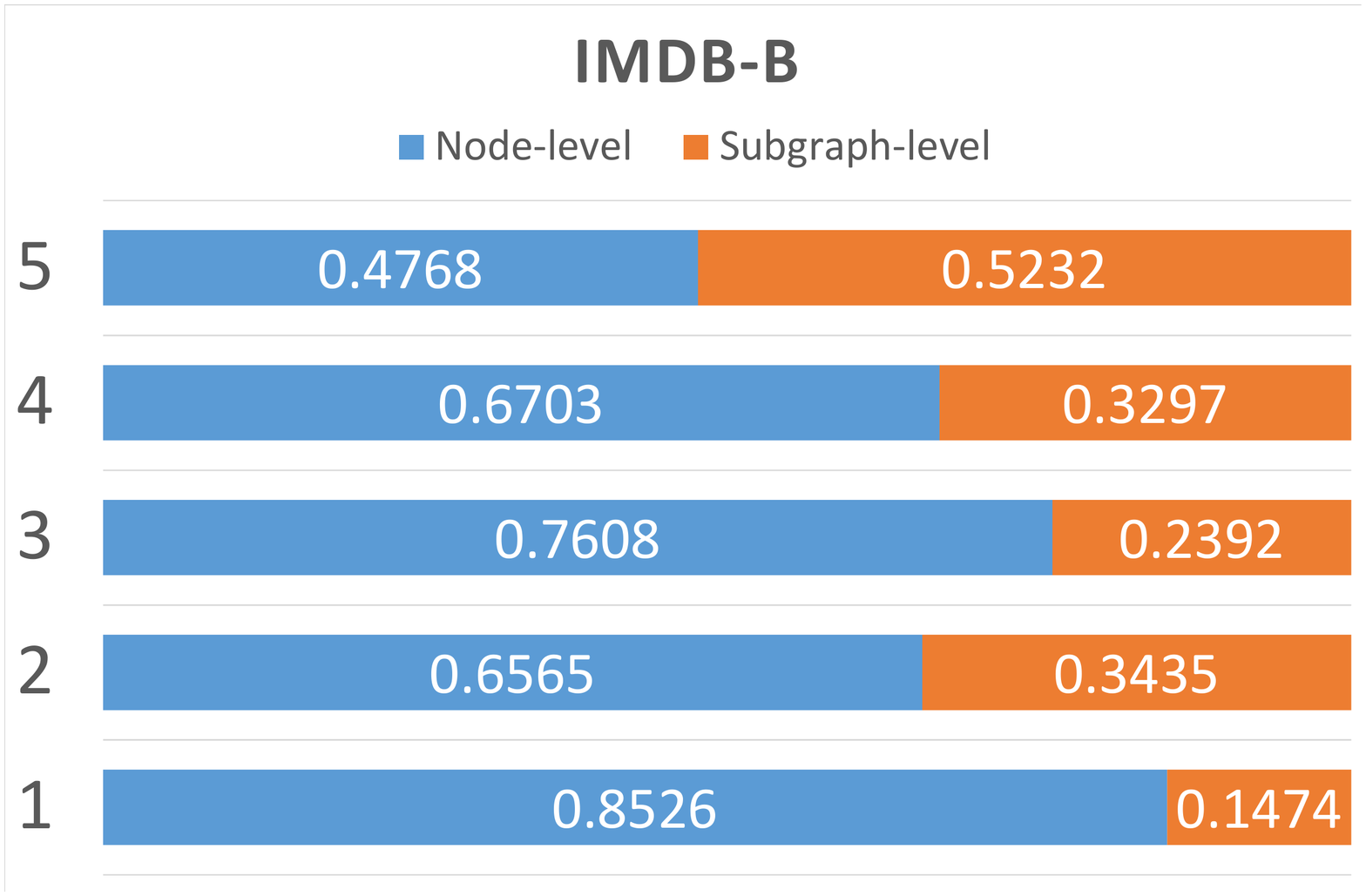}
\end{minipage}
}
\centering
\caption{Variation of the two-level attention parameters on real-world datasets. The number of histograms represents the weight coefficient.}
\label{attention weight visualization}
\end{figure*}

\textbf{Performance comparison on SPNG}

\begin{figure*}[htbp]
\centering
\subfigure[MUTAG.]{
\begin{minipage}[H]{0.3\linewidth}
\centering
\includegraphics[width=2in]{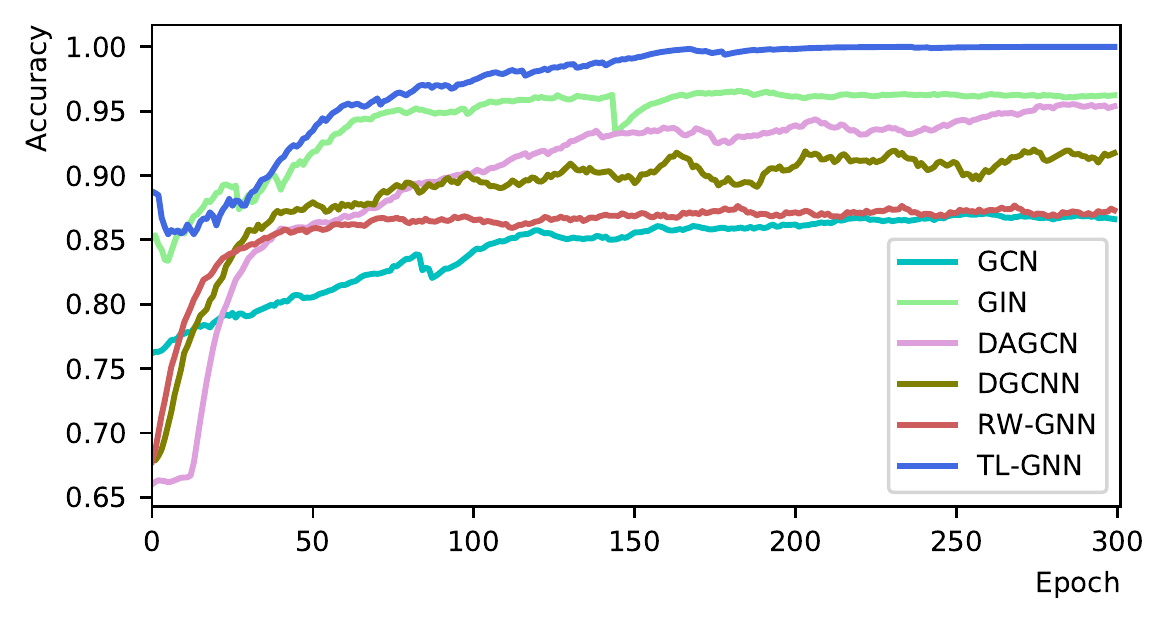}
\end{minipage}
}
\subfigure[PTC.]{
\begin{minipage}[H]{0.3\linewidth}
\centering
\includegraphics[width=2in]{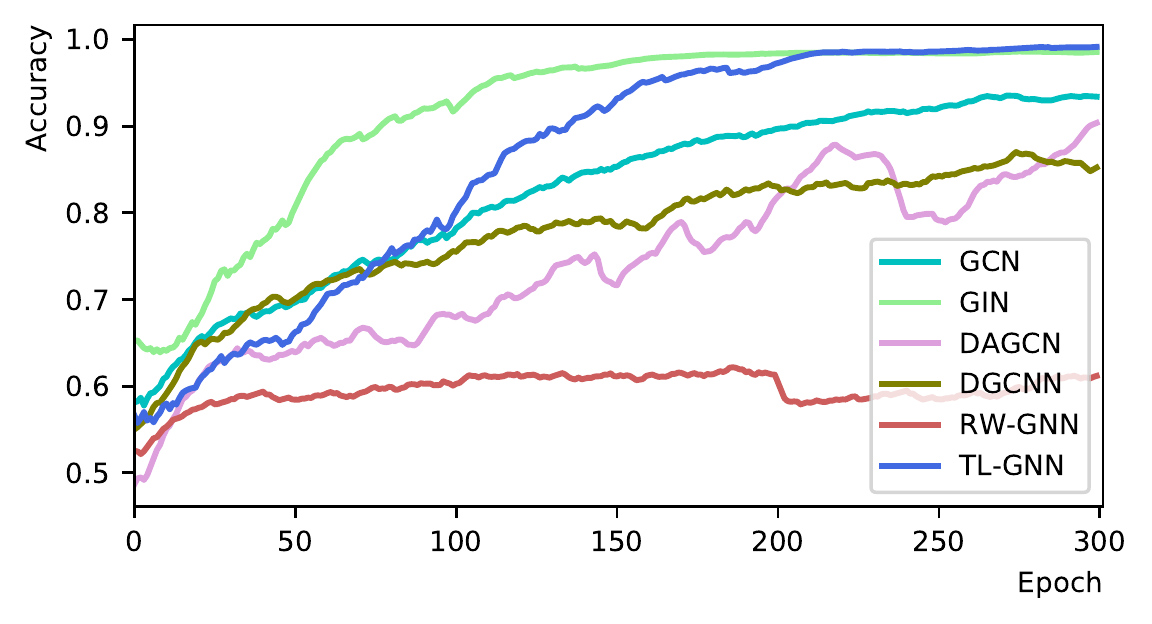}
\end{minipage}
}
\subfigure[NCI1.]{
\begin{minipage}[H]{0.3\linewidth}
\centering
\includegraphics[width=2in]{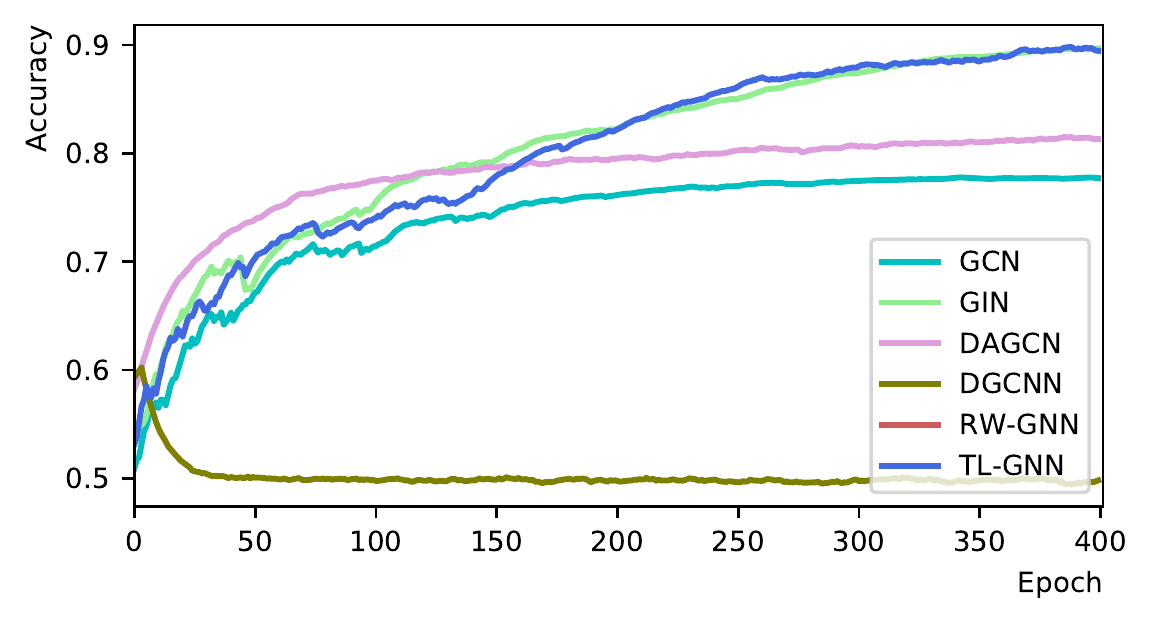}
\end{minipage}
}
\subfigure[COX2.]{
\begin{minipage}[H]{0.3\linewidth}
\centering
\includegraphics[width=2in]{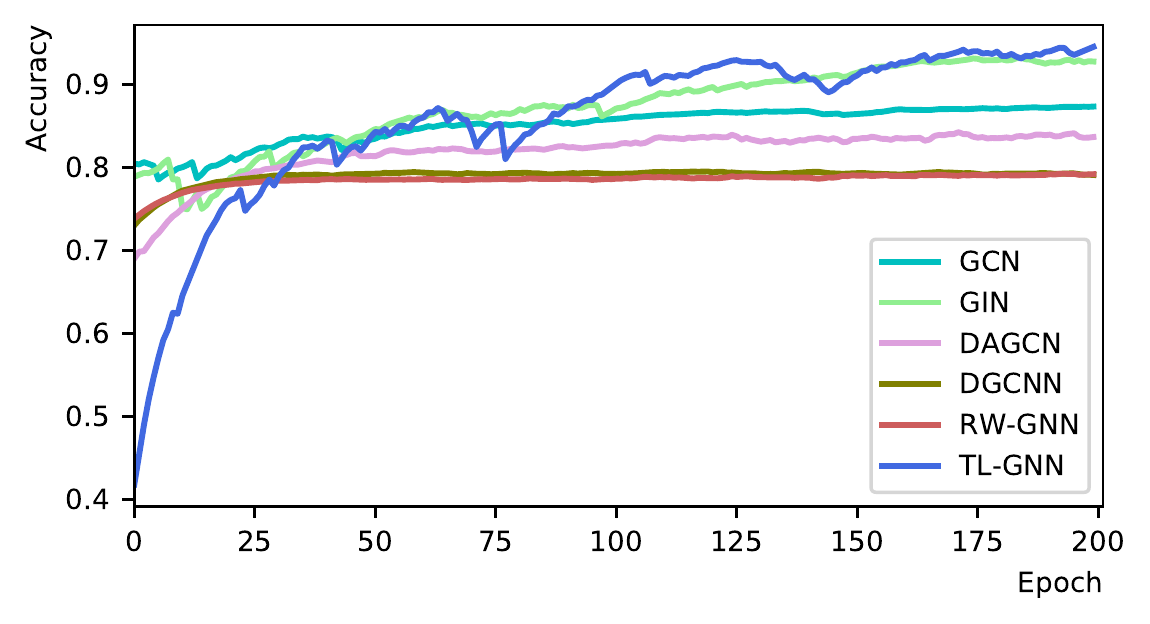}
\end{minipage}
}
\subfigure[PROTEINS.]{
\begin{minipage}[H]{0.3\linewidth}
\centering
\includegraphics[width=2in]{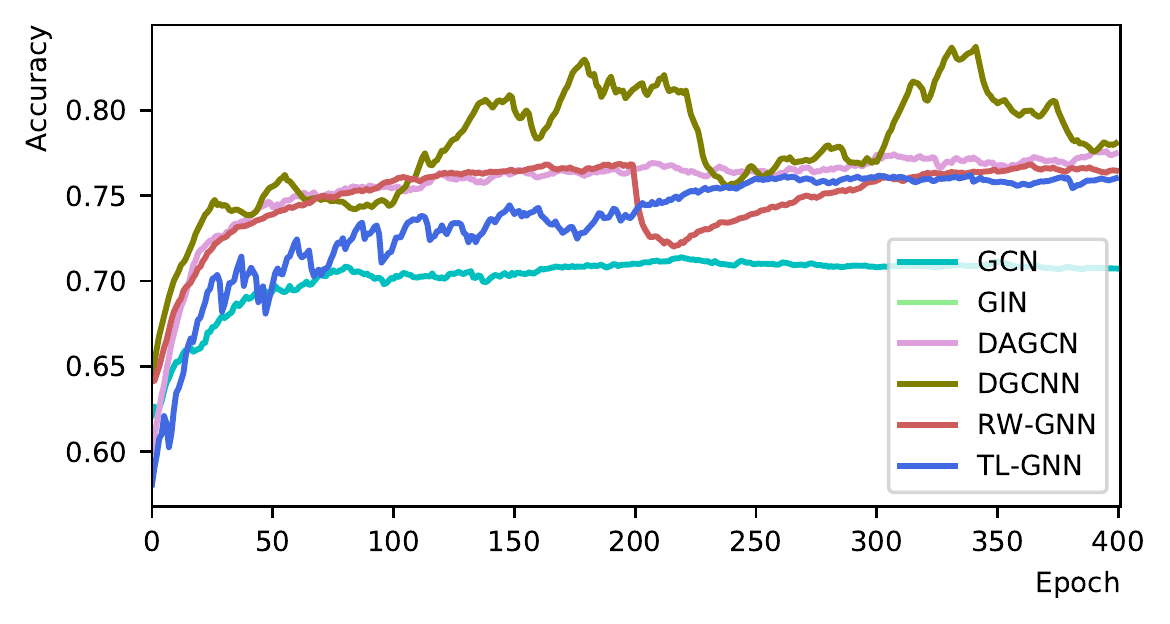}
\end{minipage}
}
\subfigure[SPNG.]{
\begin{minipage}[H]{0.3\linewidth}
\centering
\includegraphics[width=2in]{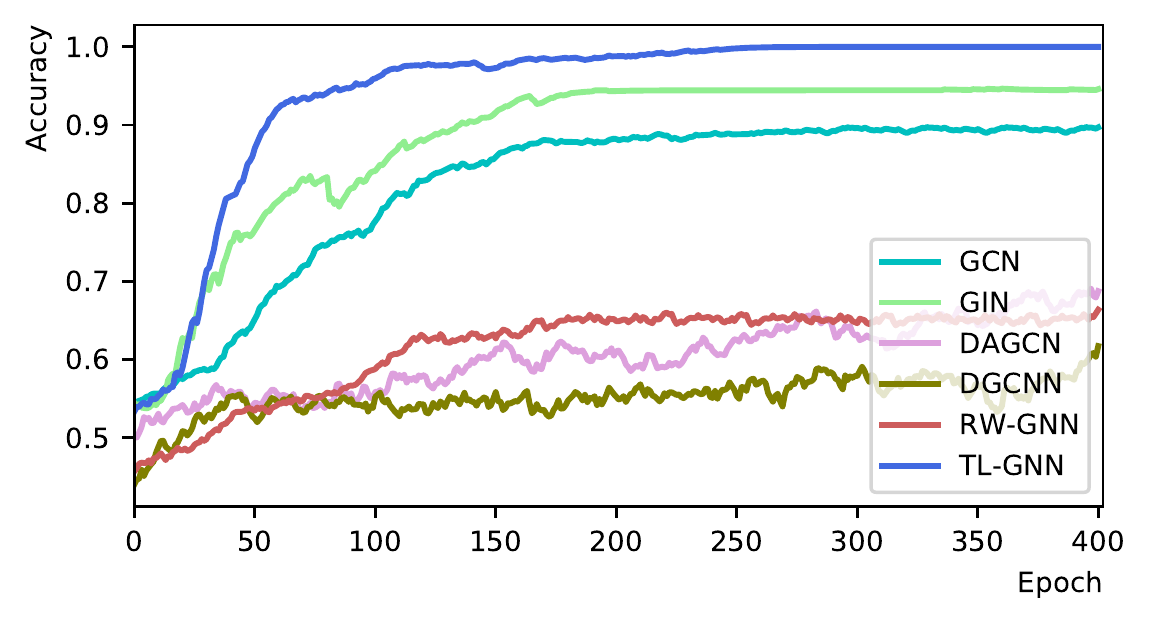}
\end{minipage}
}
\subfigure[IMDB-M.]{
\begin{minipage}[H]{0.3\linewidth}
\centering
\includegraphics[width=2in]{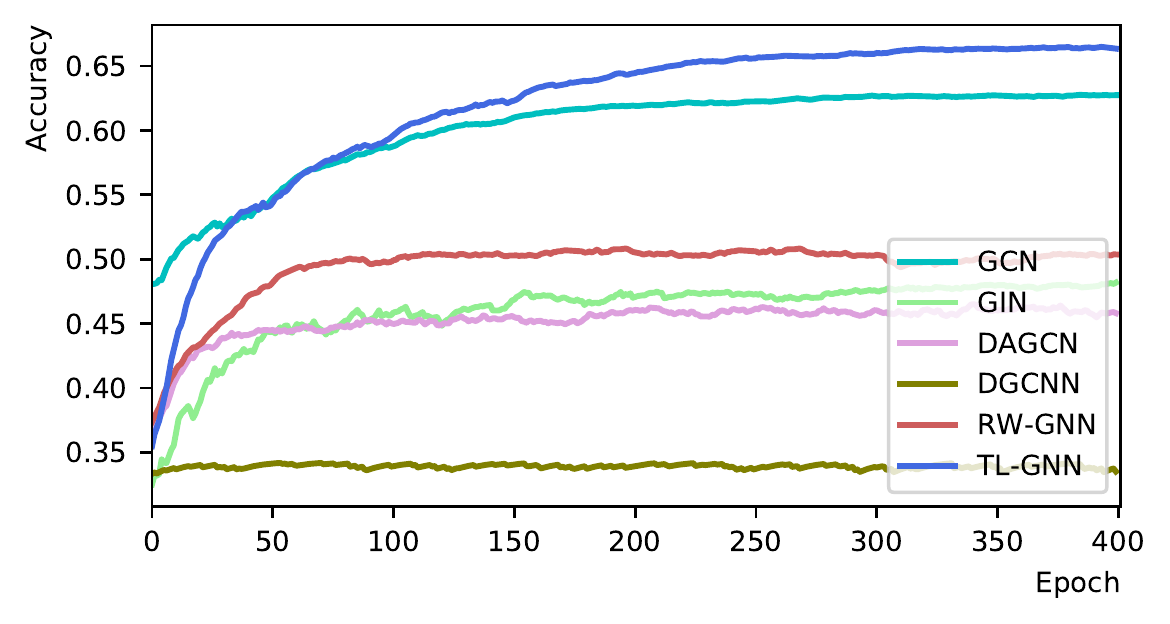}
\end{minipage}
}
\subfigure[IMDB-B.]{
\begin{minipage}[H]{0.3\linewidth}
\centering
\includegraphics[width=2in]{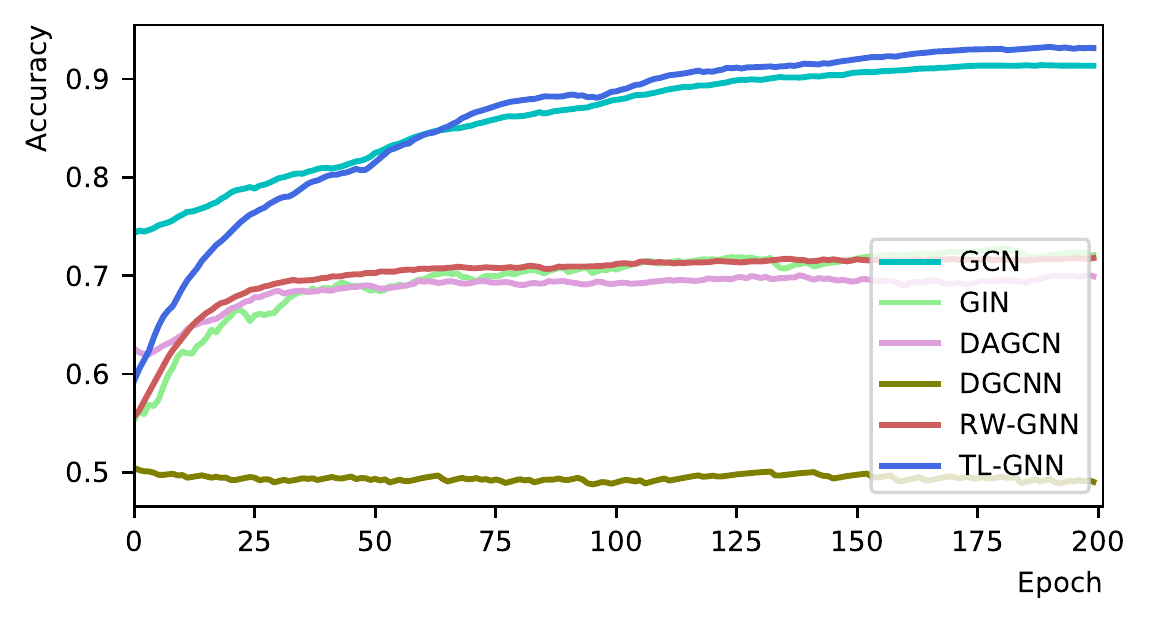}
\end{minipage}
}
\centering
\caption{Training performances comparison.}
\label{training acc}
\end{figure*}

An interesting observation that can be made from Table.~\ref{synthetic result} is that the TL-GNN easily achieves 100\% training accuracy while the alternative methods studied do not. None of the methods can perfectly fit the training data with the exception of TL-GNN. Due to the fact that both GIN and GCN with $K$ layers can capture $K-$hop information, they do offer some robustness to the LPI problem and can achieve 95.6\% and 91.1\% training accuracy respectively. However, the remaining GNN training accuracies are no more than 80\%. The results on SPNG show that the TL-GNN can distinguish PNGs perfectly. As shown in Fig.~\ref{training acc}, the TL-GNN training accuracy after 50 epochs reaches 90\%, which represents the fastest convergence rate. Finally, the TL-GNN converged at a stable 100\% training accuracy after 220 iterations. The training accuracies of RW-GNN, DGCNN, DAGCN no longer increase after 100 epochs. The reason for this is that these methods focus on capturing the local neighborhood information, which leads to less robustness to the LPI problem. 

As for GIN, its convergence rate is slower than the TL-GNN. According to the curve shown in Fig.~\ref{training acc}, as the training proceeds, GIN saturated at 95.6\%. We believe that the TL-GNN is benefited significantly from the subgraph merging strategy described in Section~\ref{Framework of TL-GNN} because both GIN and TL-GNN have the same GNN layers, but TL-GNN's overall performance is better.

\begin{table}[ht]
\centering
\caption{\centering{TRAINING ACCURACY ON SYNTEHTIC DATASET(IN \% $\pm$ STANDARD ERROR)}}
\large
\begin{tabular}{c|c}
\hline
Model &Training Accuracy \\ \hline
DGCNN& $64.0$ \\
DAGCN& $72.0$\\
GIN& $95.6$\\
GCN& $91.1$\\
RW-GCN& $70.37$\\\hline
TL-GNN& $\bm{100}$\\
\hline
\end{tabular}
\label{synthetic result}
\end{table}

\textbf{Variants of supernode-based subgraph representation}

{\color{red}{To demonstrate the effectiveness of supernode-based subgraph-level representation used by the proposed methods, we provide a similar TL-GNN architecture, TL-GNN(w/o S) that accepts separated subgraphs as the input sequentially and regards each subgraph as a separated graph to capture its feature, instead of a generated graph containing supernodes. The results of this experiment are shown in Table.~\ref{test acc}. It is obvious that TL-GNN achieves better performance than TL-GNN (w/o S) because TL-GNN(w/o S) neglects connections between subgraphs. The reason for this observation is supernode-based subgraph-level representation provides connections between subgraphs, which is beneficial to capture subgraph-level information. However, TL-GNN (w/o S) achieves better performances than other baselines on most datasets, because of the subgraph-level information captured by TL-GNN (w/o S).}}

\textbf{Variants of AGG\_SUB and MERG}

In order to verify the theory described in Section.~\ref{Framework of TL-GNN}, we select both non-injective function and injective functions for the AGG\_SUB and MERG operators and then compare their performance. We chose the sum function as the injective function. For the non-injective function, we chose the max function. Although there is a large potential choice for the non-injective function, the max is the most convenient for use with our method. The specific TL-GNN variants used are summarised in Table.~\ref{TL-GNN variances}. The performances of the four TL-GNN variants are shown in Table.~\ref{test acc}. As shown in Table.~\ref{test acc}, TL-GNN achieves a better performance than the remaining variants of TL-GNN on most of the datasets studied. For TL-GNN\_mm, on all the datasets except PROTEINS, TL-GNN\_mm is the weakest performing TL-GNN variant. This observation demonstrates the correctness of Lemma 6. When AGG\_SUB and MERG are not injective functions, the TL-GNN can not effectively capture and retain subgraph-level information. For TL-GNN\_ms and TL-GNN\_sm, the performance is better than TL-GNN\_mm but poorer than TL-GNN. We believe this is because a non-injective function leads to information loss.

\begin{table}[ht]\footnotesize
\centering
\caption{\centering{DETAILED CONFIGURATION OF TL-GNN variances}}
\normalsize
\begin{tabular}{|c|c|c|}\hline
TL-GNN variances & AGG\_SUB &MERG\\ \hline
TL-GNN& sum& sum\\ \hline
TL-GNN\_sm& sum& max\\ \hline
TL-GNN\_ms& max& sum\\ \hline
TL-GNN\_mm& max& max\\ \hline
\end{tabular}
\label{TL-GNN variances}
\end{table}

\textbf{Attention weight visualization and analysis}

{\color{red}{We provide visualization results of each layer on each dataset in Fig.~\ref{attention weight visualization}. The blue and orange represent attention weights of node-level and subgraph-level respectively. Numbers of the vertical axis are layers. We applied 3 layers on MUTAG, PTC, COX2, SPNG and 5 layers on NCI1, PROTEINS, IMDB-B, IMDB-M, PROTEINS. An observation is that TL-GNN pays more attention to node-level on all datasets. However, in most cases, the attention weights of subgraph-level increased with the increasing of the layer. A possible reason is that node features become identical with the layer increasing, namely the oversmoothing problem. So, the TL-GNN pays more attention to subgraph-level in high layers. The ratios of node-level attention weight to subgraph-level attention weight are around 7:3. This observation shows that node-level information is more important to the graph classification task and little subgraph-level information is beneficial to graph classification.}}

\begin{figure*}[htbp]
\centering
\subfigure[Training time.]{
\begin{minipage}[H]{0.4\linewidth}
\centering
\includegraphics[width=2.5in]{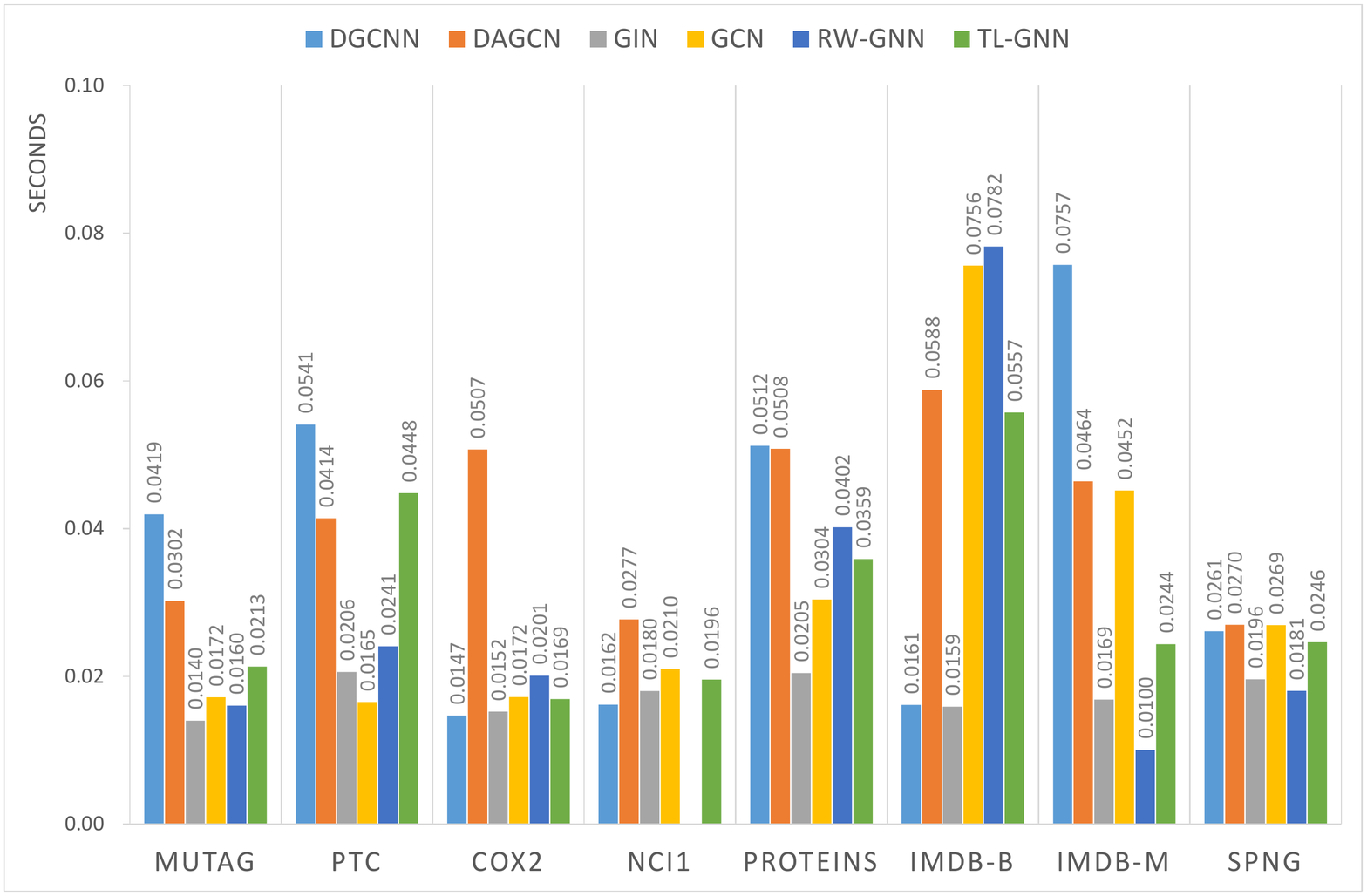}
\end{minipage}
}
\subfigure[Test time.]{
\begin{minipage}[H]{0.4\linewidth}
\centering
\includegraphics[width=2.5in]{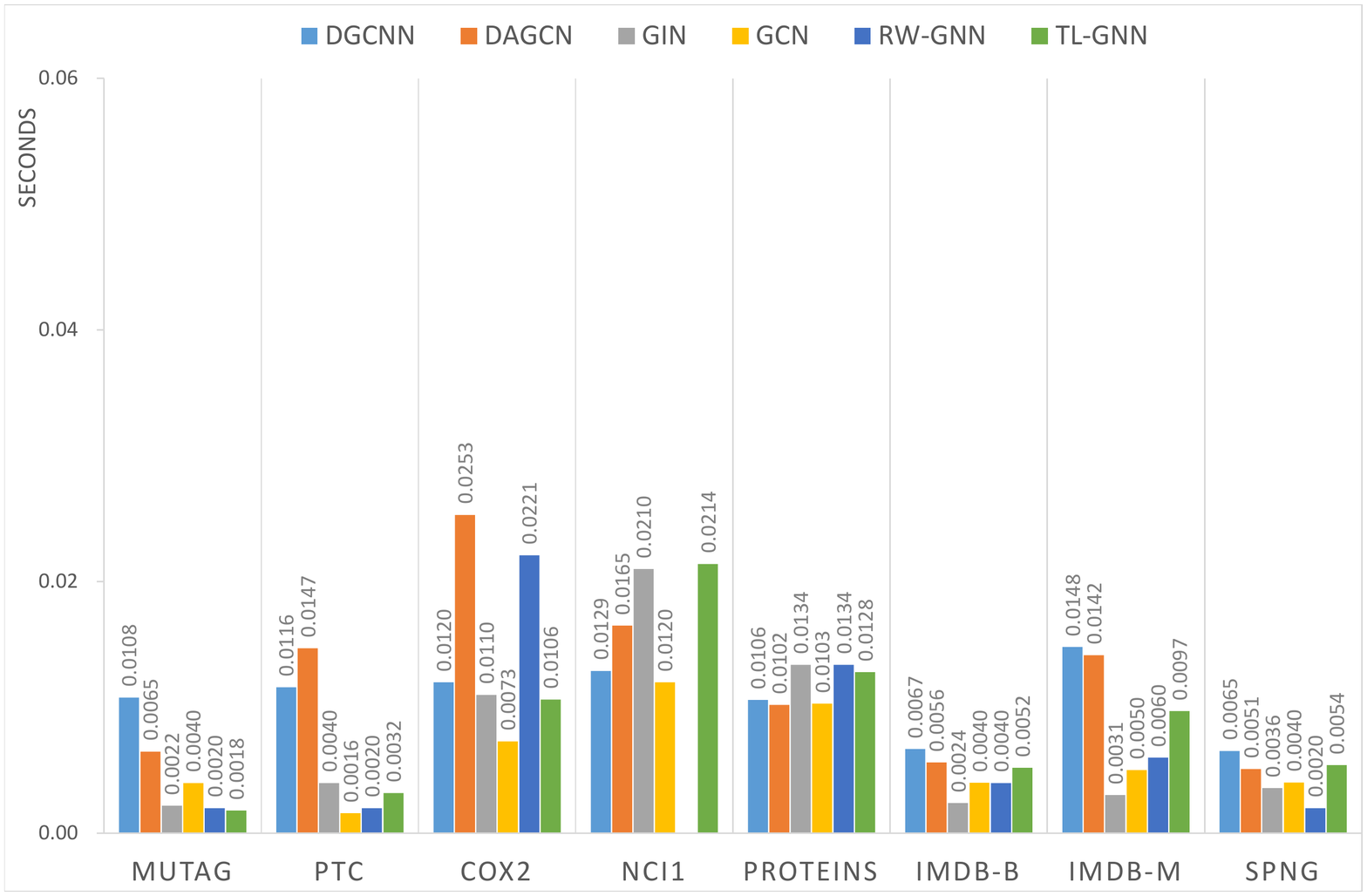}
\end{minipage}
}
\centering
\caption{Running time comparison.}
\label{running time}
\end{figure*}

\begin{figure*}[htbp]
\centering
\subfigure[MUTAG.]{
\begin{minipage}[H]{0.23\linewidth}
\centering
\includegraphics[width=1.9in]{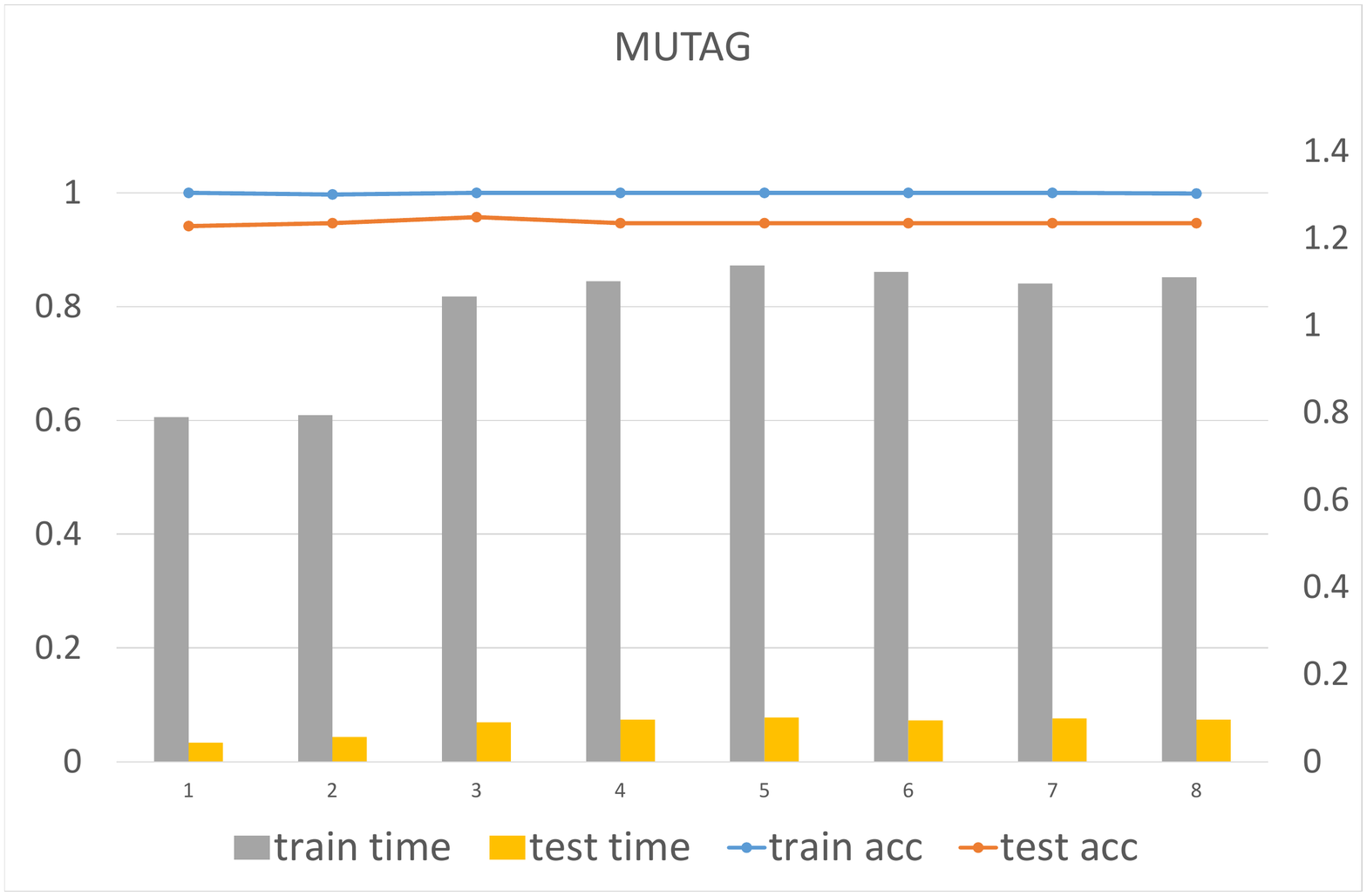}
\end{minipage}
}
\subfigure[PTC.]{
\begin{minipage}[H]{0.23\linewidth}
\centering
\includegraphics[width=1.9in]{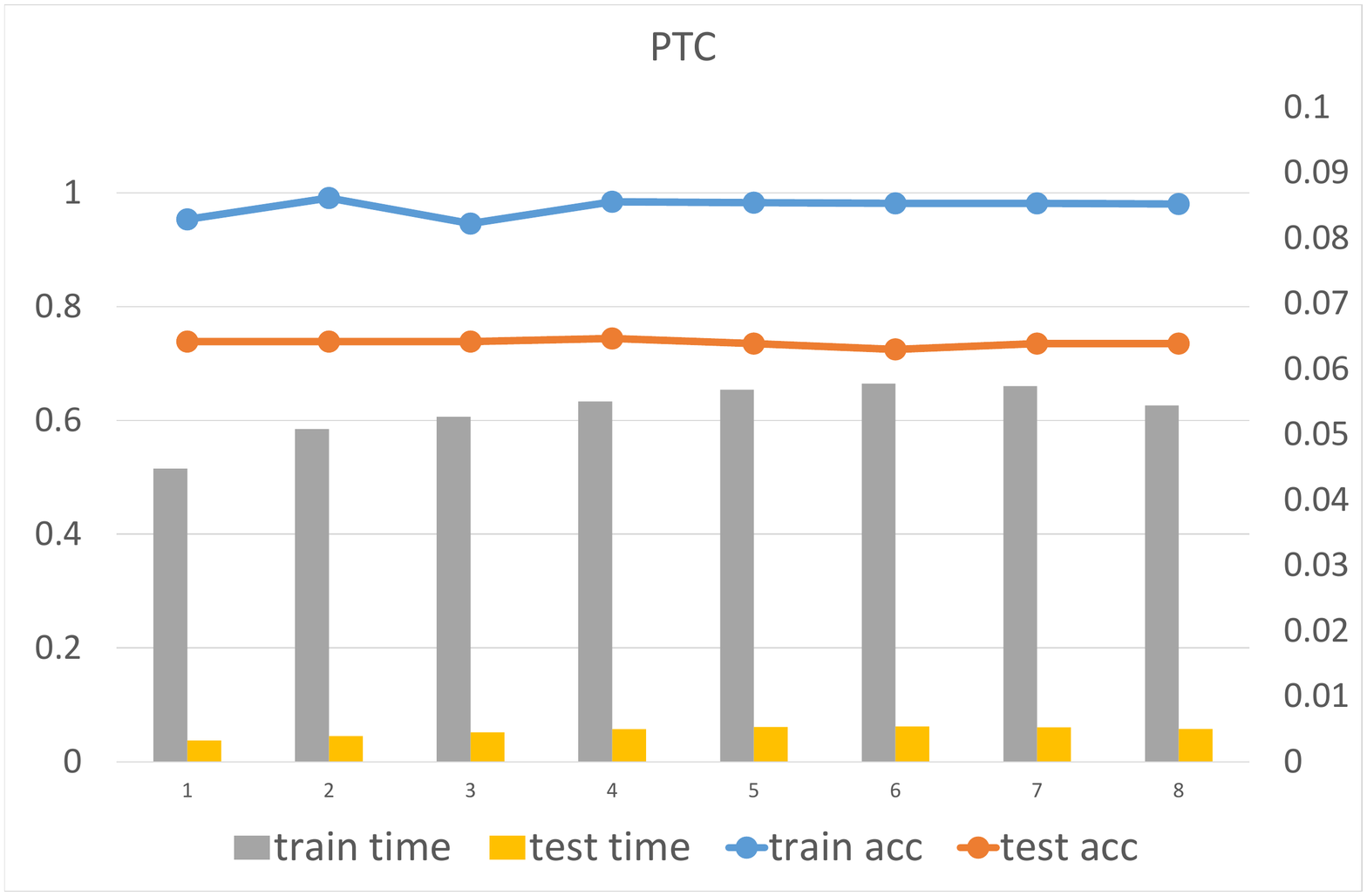}
\end{minipage}
}
\subfigure[NCI1.]{
\begin{minipage}[H]{0.23\linewidth}
\centering
\includegraphics[width=1.9in]{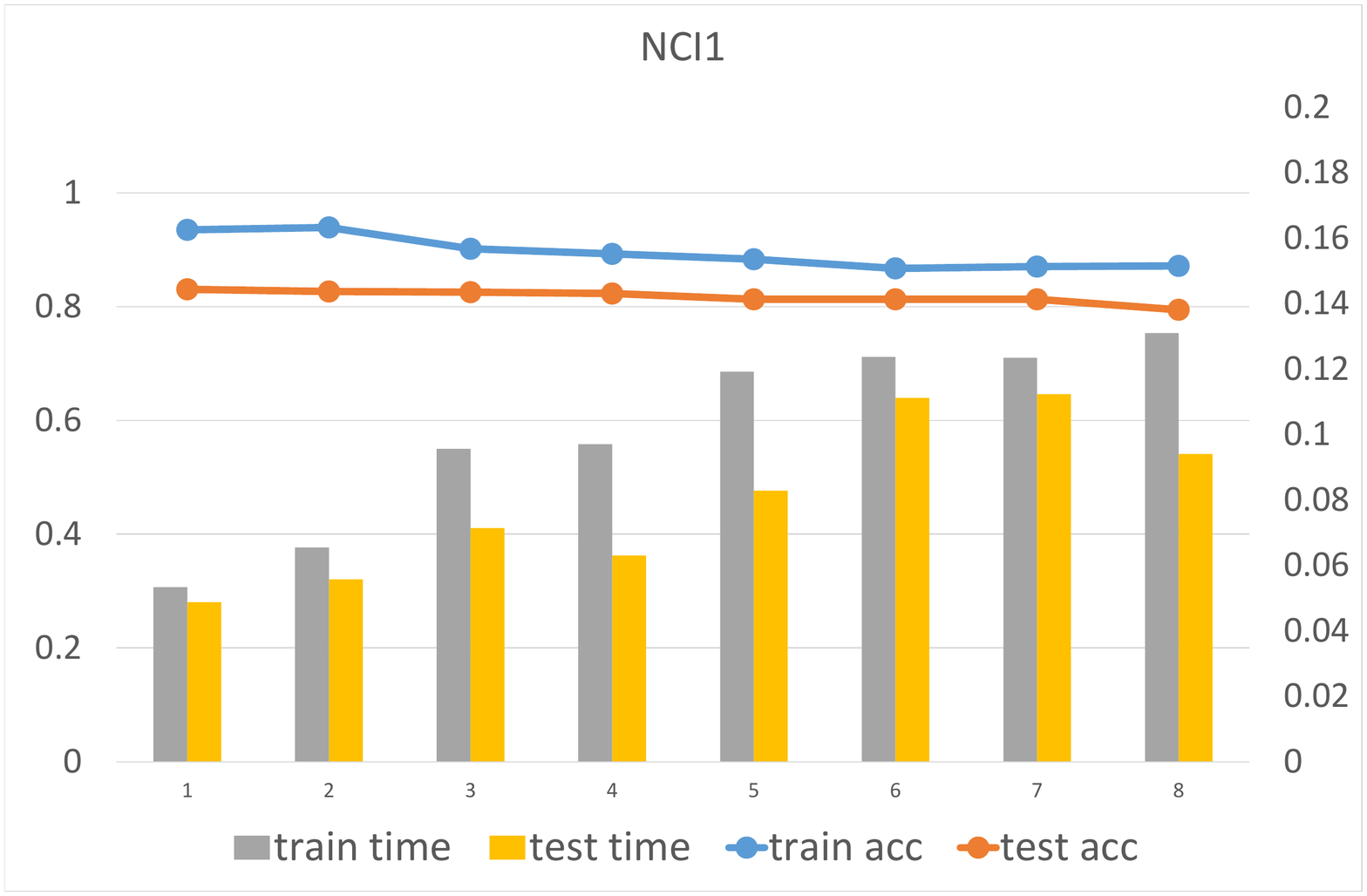}
\end{minipage}
}
\subfigure[COX2.]{
\begin{minipage}[H]{0.23\linewidth}
\centering
\includegraphics[width=1.9in]{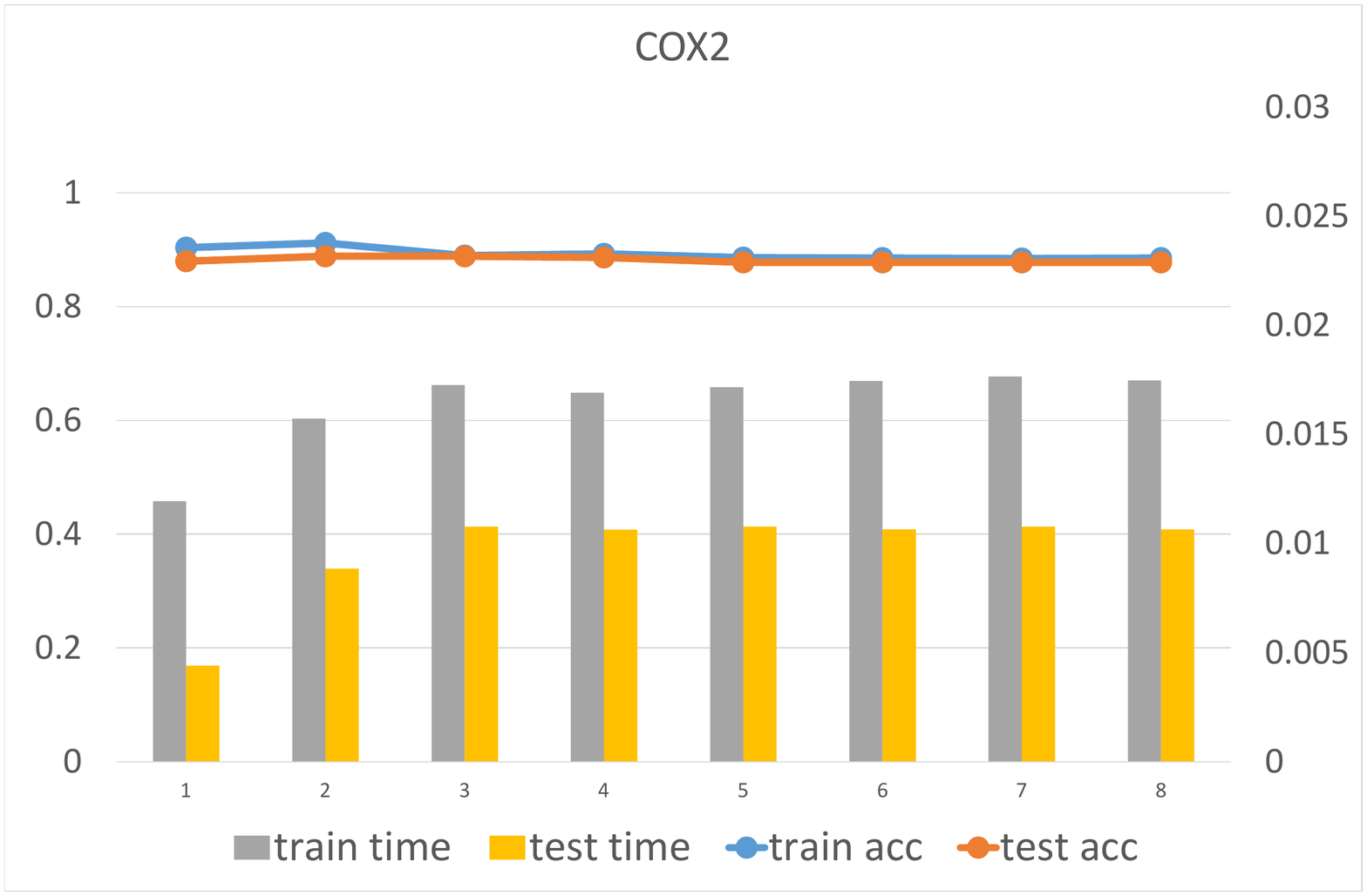}
\end{minipage}
}
\subfigure[IMDB-M.]{
\begin{minipage}[H]{0.23\linewidth}
\centering
\includegraphics[width=1.9in]{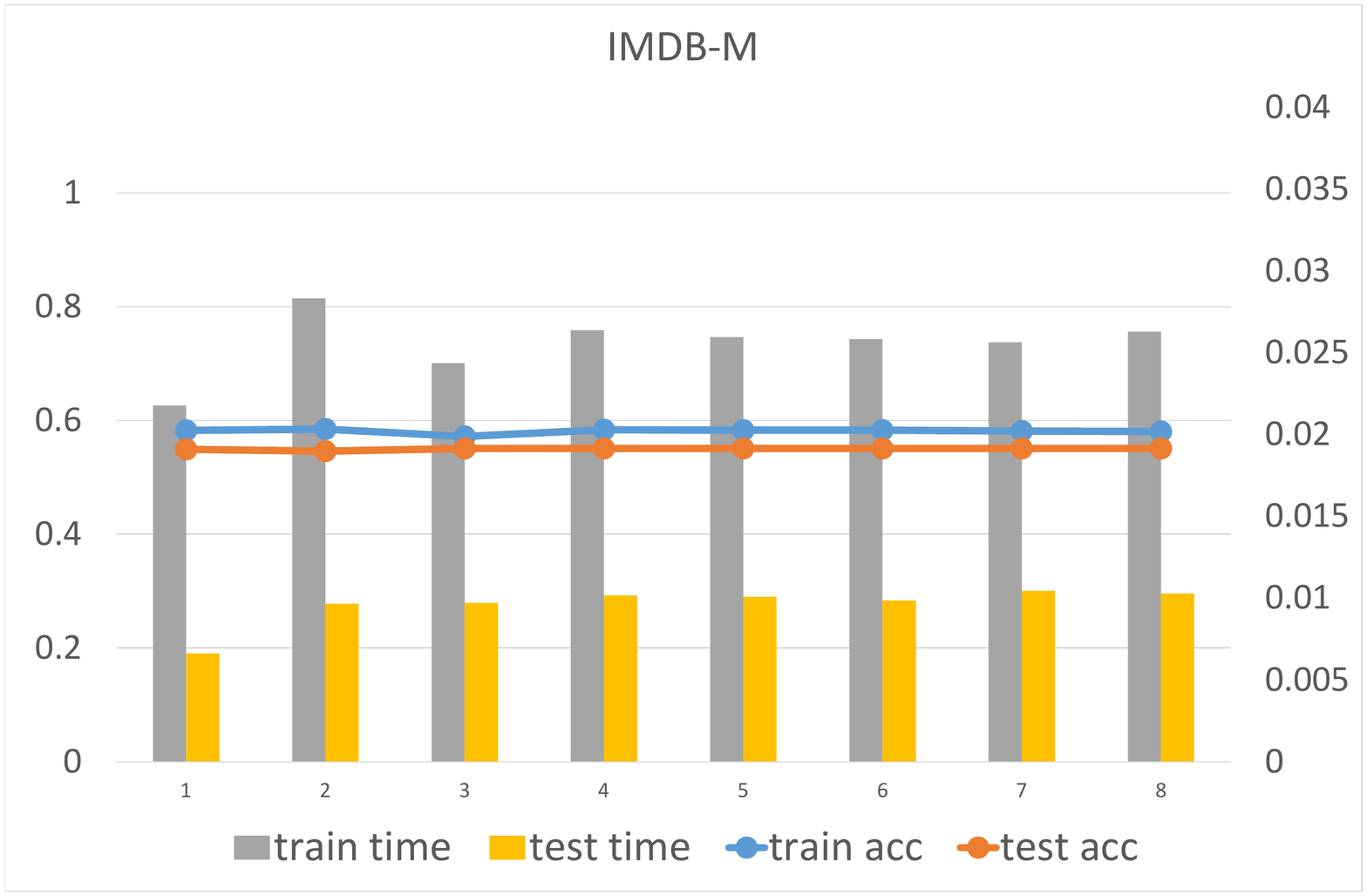}
\end{minipage}
}
\subfigure[IMDB-B.]{
\begin{minipage}[H]{0.23\linewidth}
\centering
\includegraphics[width=1.9in]{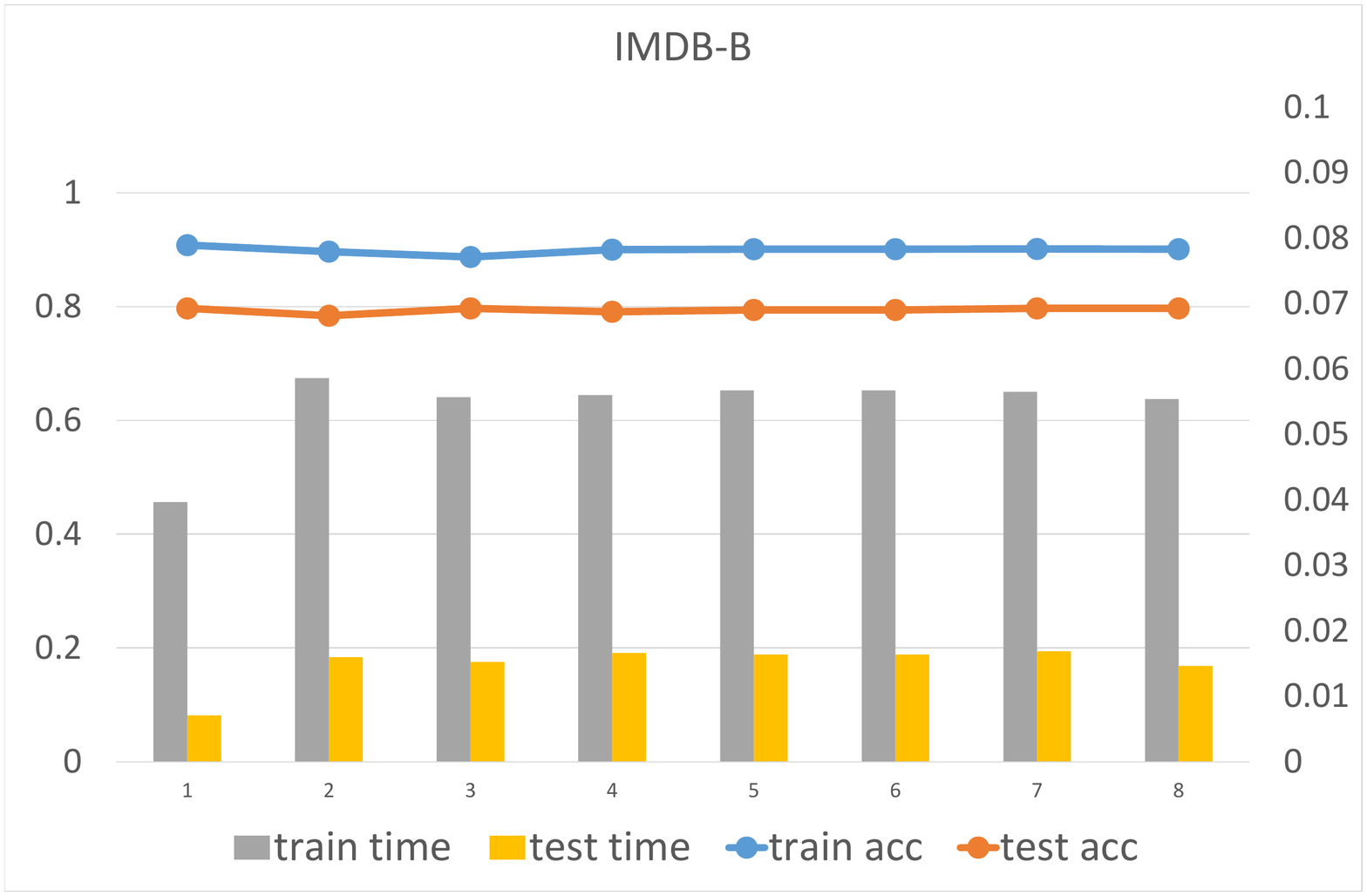}
\end{minipage}
}
\subfigure[PROTEINS.]{
\begin{minipage}[H]{0.23\linewidth}
\centering
\includegraphics[width=1.9in]{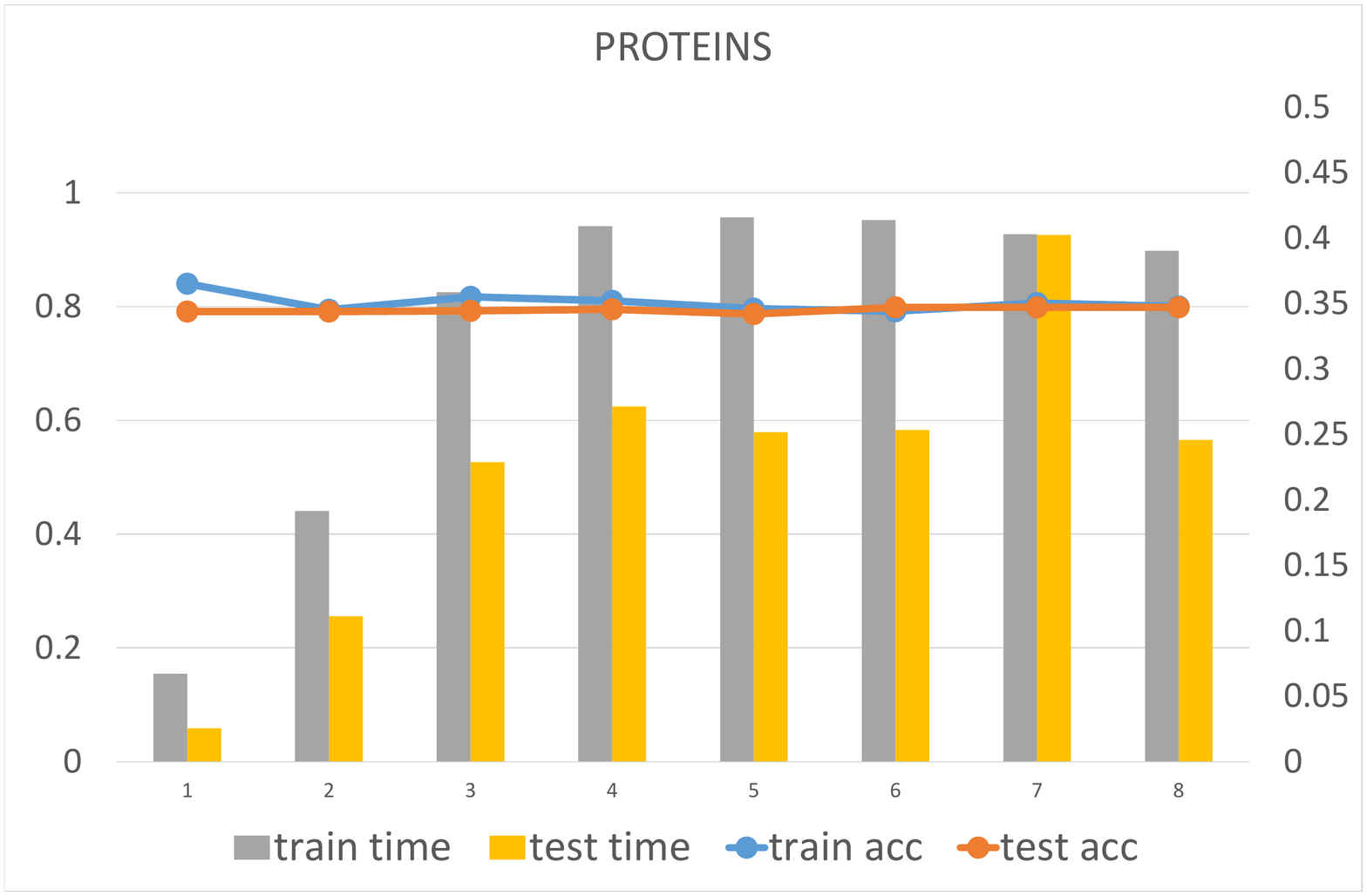}
\end{minipage}
}
\centering
\caption{Variation of the parameter $D$ on real-world datasets. The horizontal axis shows number of $D$.}
\label{varify scale}
\end{figure*}

\textbf{Comparison of computational complexity}

{\color{red}{We report the average running times per iteration after training or test of TL-GNN and the baselines Fig.~\ref{running time}. For a fair comparison, all of the methods are run on a system with an Intel Xeon CPU E3-1270 v5 processor. Due to RW-GNN cannot accept NCI1 as input\cite{nikolentzos2020random}, the running time of RW-GNN on NCI1 dataset is vacant. Obviously, TL-GNN requires more time than GIN because of the extra computation of subgraph-level and attention mechanisms. Although GIN and TL-GNN have identical GNN layers, TL-GNN costs more time on both training and test than GIN. The main reason is that TL requires additional subgraph convolution operations. However, the training and test time of TL-GNN is less than several baselines such as DGCNN and RWGNN, because of the simple but effective architecture of TL-GNN.}}

\textbf{Variation of the depth}

In order to find the optimal value of $D$, we have carried out experiments on all of the real-world datasets, and compute time and accuracies for both training and test are presented. From Fig.\ref{varify scale} the compute time consumed in training and testing increases with an increasing value of $D$ on most of the datasets studied. A large value of $D$ leads to a large number of subgraphs and thus increases the compute time required.

When $D$ reaches a critical value, the compute time consumed in both training and testing ceases to increase. This is because when $D$ is large enough, then $2^D$ is greater than the number of nodes in the graph. This means that all subgraphs within a graph have been located. For example, on the MUTAG dataset, the training becomes stable when $D$ is greater than 5, and this is because the mean node degree of MUTAG is less than 18. For most of the datasets studied, the maximum test accuracy is reached when $D$ is 3 or 4. This indicates that aggregating the subgraph information for the 8-hop or 16-hop neighbors of a node adds the greatest benefits to the graph classification task. 
It is clear that too small a $D$ value leads to an insufficient number of subgraphs, thus TL-GNN can not capture the subgraph-level information well. Conversely, too large a value of $D$ leads to subgraphs that are distant from the node in question being included. The correlation between these subgraphs and the node in question is low, and this is not conducive to improved performance.

\textbf{Comparison of training performance}

We compare the effect of training for several GNNs for which the authors have made their code available. This study is summarised in Fig.~\ref{training acc}. From the table, it is clear that TL-GNN can better fit all of the datasets studied than the alternative GNNs. For example, TL-GNN is the only method which achieves 100\% training accuracy on MUTAG. The same observation applies to SPNG. As for the NCI1 and PTC datasets, GIN and TL-GNN are the best performing methods, and their training accuracies are roughly identical. On PTC, the convergence speed of TL-GNN is slower than that of GIN, but faster than that of the remaining methods. On the COX2 dataset, the convergence rates of TL-GNN and GIN are close, but the final training accuracy of TL-GNN is slightly higher than that of GIN. On the PROTEINS dataset, the training accuracy of TL-GNN does not outperform that of DGCNN, which is equal to that of the other methods studied. On the two variants of IMDB, TL-GNN and GCN achieve the best training performance. However, TL-GNN is slightly, but consistently better than GCN. We also observe that TL-GNN achieves a higher training accuracy than GIN on many of the datasets studied. Due to the fact that they have the same GNN layer structure, these improvements come from the richer sources of information exploited by TL-GNN. It is worth noting that the convergence speed of TL-GNN on the SPNG dataset is much faster than the remaining deep learning methods. The above observations demonstrate that the subgraph-level information captured by TL-GNN is not only helpful in distinguishing PNGs, but also beneficially enriches the graph representation.

\section{Conclusion}
\label{Conclusion}
In this paper, we presented a novel deep learning method for graph-structured data, the so-called Two-level Graph Neural Network (TL-GNN). Considering the representational limitations on GNNs caused by the LPI problem, we introduce subgraph-level information into our framework and propose two novel operators, AGG\_SUB and MERG to implement our method. Moreover, we provide distinct mathematical definitions for permutation non-isomorphic graphs (PNGs) and also provide a theoretical analysis of the role of subgraphs in solving the LPI problem. A novel subgraph counting algorithm is also proposed, which can locate all subgraphs of a $n-$node graph within a $D-$hop neighborhood of each node; this has $O(Dn^3)$ time complexity and $O(Dn^3)$ space complexity. Experimental results show that TL-GNN achieves state-of-the-art performance on most real-world datasets and distinguishes all the data perfectly on the synthetic PNG dataset.
As for further work, we plan to extend the method to heterogeneous graphs and further enrich the macroscopic information captured by GNNs. Specifically, we will treat the different types of subgraphs as heterogeneous nodes within a heterogeneous graph and then exploit a discriminative attention mechanism to differentiate between these nodes. Besides, we will further discuss and explore the influence of the LPI problem on the node classification task.

\ifCLASSOPTIONcaptionsoff
  \newpage
\fi


\bibliographystyle{IEEEtran}
\bibliography{reference.bib}


%

\end{document}